\documentclass{article}

\usepackage{graphicx}
\usepackage{amsmath}
\usepackage{amssymb}
\usepackage{booktabs}
\usepackage{lipsum}  
\usepackage{color}
\usepackage{xcolor}
\usepackage{multirow}
\usepackage{multicol}

\usepackage{microtype}

\usepackage{xparse}

\ExplSyntaxOn
\cs_new:Nn\__avercalc_plus:n{
  + ( #1 )
}
\NewExpandableDocumentCommand{\avercalc}{O{1}+m}{%
  \fp_eval:n {
    round(
      ( 0 \clist_map_function:nN { #2 } \__avercalc_plus:n ) / max(1, \clist_count:n { #2 })
      , #1
    )
  }
}
\ExplSyntaxOff

\DeclareMathAlphabet\mathbfcal{OMS}{cmsy}{b}{n}

\usepackage{array}
\newcolumntype{H}{>{\setbox0=\hbox\bgroup}c<{\egroup}@{}}

\usepackage[pagebackref,breaklinks=true,colorlinks,citecolor=citecolor,bookmarks=false]{hyperref}

\usepackage[accepted]{icml2023}

\usepackage{mathtools}
\usepackage{amsthm}

\usepackage[capitalize,noabbrev]{cleveref}

\newcommand{\method}{\mbox{NOVIS}}

\theoremstyle{plain}

\theoremstyle{definition}

\theoremstyle{remark}

\usepackage{xspace}

\makeatletter
\DeclareRobustCommand\onedot{\futurelet\@let@token\@onedot}
\def\@onedot{\ifx\@let@token.\else.\null\fi\xspace}

\def\eg{\emph{e.g}\onedot} 
\def\ie{\emph{i.e}\onedot}

\makeatother
\usepackage{subcaption}

\icmltitlerunning{\method{}: A Case for End-to-End Near-Online Video Instance Segmentation}

\begin{document}

\twocolumn[
    \icmltitle{\method{}: A Case for End-to-End Near-Online \\ Video Instance Segmentation}
    %
    %
    %
    \icmlsetsymbol{equal}{*}
    \begin{icmlauthorlist}

    \icmlauthor{Tim Meinhardt}{tum,nvidia}
    \icmlauthor{Matt Feiszli}{meta}
    \icmlauthor{Yuchen Fan}{meta}
    \icmlauthor{Laura Leal-Taixé}{tum,nvidia}
    \icmlauthor{Rakesh Ranjan}{meta}
    \end{icmlauthorlist}
    \icmlaffiliation{tum}{Technical University of Munich}
    \icmlaffiliation{meta}{Meta Inc.}
    \icmlaffiliation{nvidia}{NVIDIA}
    \icmlcorrespondingauthor{Tim Meinhardt}{tim.meinhardt@tum.de}
    %
    \icmlkeywords{video object segmentation, near-online, transformers}
    \vskip 0.3in
]



\printAffiliationsAndNotice{}  

\begin{abstract}
    Until recently, the Video Instance Segmentation (VIS) community operated under the common belief that offline methods are generally superior to a frame by frame online processing.
    However, the recent success of online methods questions this belief, in particular, for challenging and long video sequences.
    %
    %
    We understand this work as a rebuttal of those recent observations and an appeal to the community to focus on dedicated near-online VIS approaches.
    %
    %
    %
    %
    To support our argument, we present a detailed analysis on different processing paradigms and the new end-to-end trainable \method{} (Near-Online Video Instance Segmentation) method.
    Our transformer-based model directly predicts spatio-temporal mask volumes for clips of frames and performs instance tracking between clips via overlap embeddings.
    \method{} represents the first near-online VIS approach which avoids any handcrafted tracking heuristics.
    %
    %
   %
   We outperform all existing VIS methods by large margins and provide new state-of-the-art results on both YouTube-VIS (2019/2021) and the OVIS benchmarks.
   %
\end{abstract}
\section{Introduction}
\label{sec:intro}

The computer vision community directs much of its attention to building robust video and scene understanding models.
Teaching machines to mimic or even surpass human capabilities opens vast possibilities for applications in robotics and automation such as autonomous driving.
%
%
This work focuses on the Video Instance Segmentation (VIS) task, which performs pixel level segmentation and tracking of object instances within a given RGB video input.
%
%
%
\begin{figure*}[t]
    \centering
     \begin{subfigure}[t]{0.32\textwidth}
      \includegraphics[width=\textwidth]{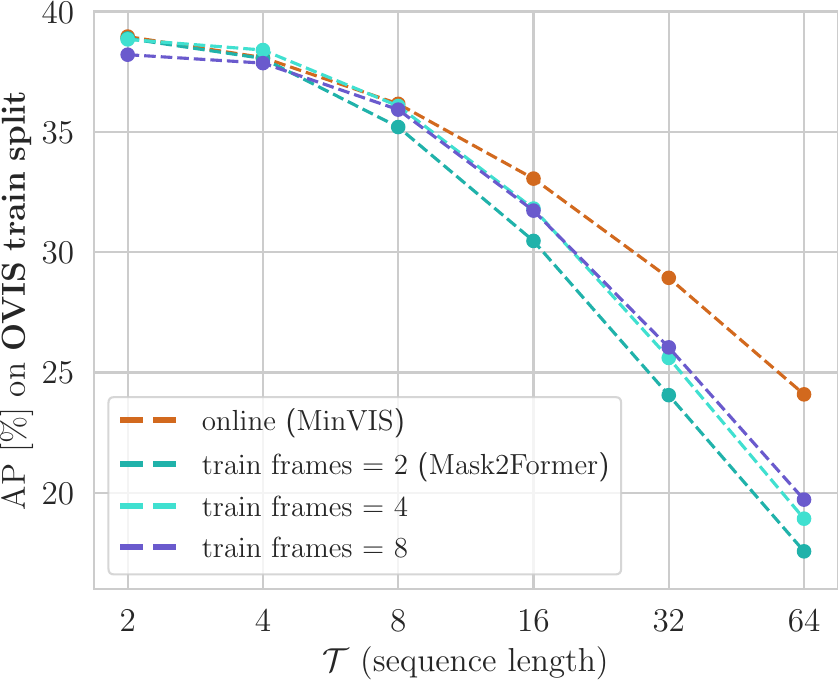}
      \caption{Offline frame generalization}
      \label{fig:teaser_offline}
    \end{subfigure}
    \begin{subfigure}[t]{0.32\textwidth}
      \includegraphics[width=\textwidth]{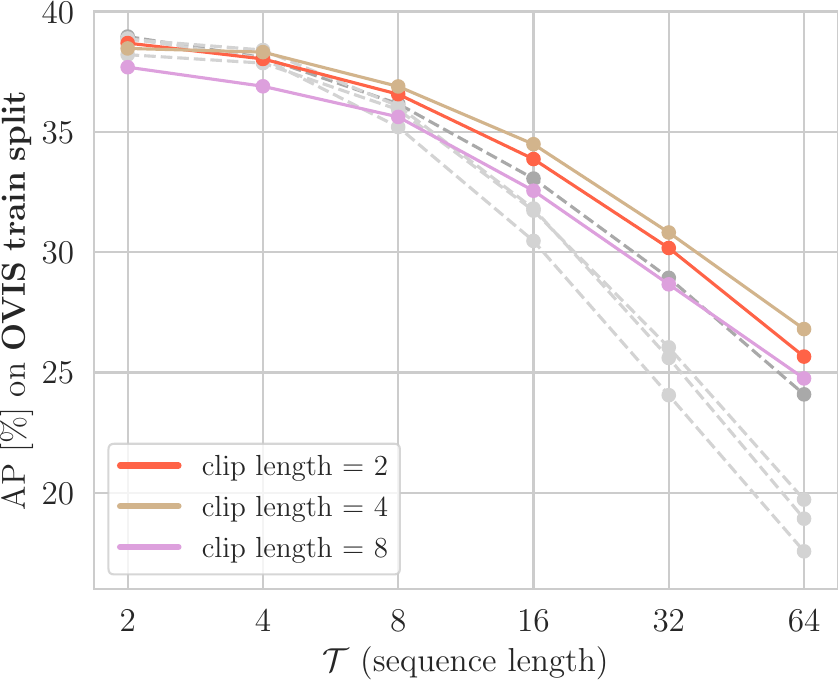}
      \caption{Near-online processing}
      \label{fig:teaser_near_online}
    \end{subfigure}
    \begin{subfigure}[t]{0.335\textwidth}
      \includegraphics[width=\textwidth]{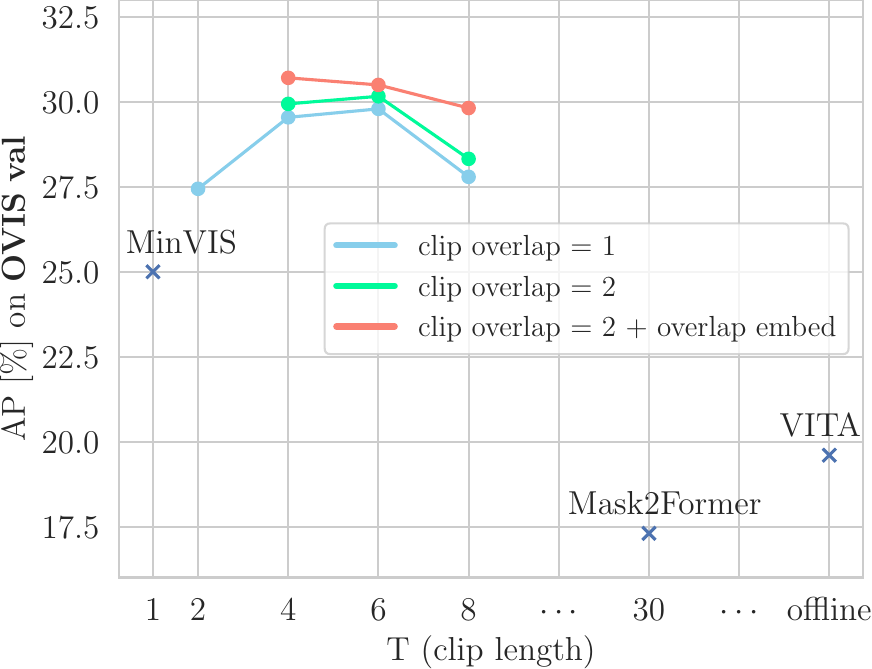}
      \caption{\textbf{\method{}} with overlap embeddings}
      \label{fig:teaser_novis}
    \end{subfigure}
    \caption{
    A \textbf{frame processing analysis} and comparison of our near-online \method{} approach with comparable online and offline baselines.
    %
    We evaluate on a 2-fold training and the official validation split of OVIS.
    Figure (a) illustrates the training to test frame generalization gap of offline methods and the superiority of online methods recently observed by the VIS community.
    The near-online variants shown in (b) do not suffer from the gap and outperform online baseline.
    In Figure (c), we demonstrate how \method{} outperforms online and offline approaches and benefits from computing \emph{overlap embeddings}.
    %
    %
    }
    \label{fig:teaser}
\end{figure*}
Since its introduction in~\cite{Yang2019vis}, the general consensus has been that the complexity of the VIS task justifies offline processing of a given video sequence.
Offline methods~\cite{stem_seg,mask_prop,prop_reduce,stmask,sg_net,seqformer,IFC,heo2022vita,mask2former4vis} process the entire sequence at once and are hence expected to provide more consistent track management and mask quality over time.
This is usually achieved by training models end-to-end on multiple frames at once.
However, the training as well as inference comes with considerable computational costs, which scale with the sequence length and often render an offline processing of long sequences infeasible.
More recently, online~\cite{IDOL, huang2022minvis} approaches, which segment and track instances on a frame-by-frame basis, have shown great potential, even surpassing existing offline methods.
%
These methods usually exhibit a smaller computational footprint which allows for the application on real-time data streams.
However, the limited temporal information (predictions must not include future information) makes occlusion handling and consistent mask predictions far more challenging.
To mitigate these issues, most online methods rely on handcrafted tracking heuristic.
%
Near-online methods~\cite{mask_prop,stem_seg,devis,IFC}, also referred to as sliding window, process clips of frames and use heuristic instance tracking between clips.
%
They combine the best of the online and offline paradigms, while avoiding their respective downsides.
%
%
Processing multiple frames at once improves mask quality and track management, and clip overlaps facilitate the clip-to-clip tracking substantially.
%

\noindent \textbf{The flawed case against (existing) offline methods.}
The original VIS benchmark only focused on short sequences which allowed offline methods to excel.
%
More recent benchmarks such as OVIS~\cite{ovis} include very long sequences that most offline methods are not able to process.
This limitation alone could explain the recent resurgence of online approaches for VIS.
However, the analysis of the IDOL~\cite{IDOL} paper goes a step further by questioning the general efficiency of existing offline methods.
In a series of oracle experiments, they show diminishing performance with increasing sequence length for offline as well as online processing.
%
However, the performance drop for offline methods~\cite{seqformer, IFC} is larger than for comparable online approaches.
%
This observation runs against the general intuition that processing all frames at once yields more robust mask predictions.
%
%
%
%
%
%
The IDOL analysis and benchmark results presented in~\cite{IDOL, huang2022minvis} represent a broader argument against the efficiency of existing offline methods advocating for a frame-to-frame online processing.
%
%
%
In this work, we draw a different conclusion by identifying the true reason for their suboptimal performance: The generalization gap between the number of train and test frames.

\noindent \textbf{Why near-online is the future of VIS.}
%
%
In~\cref{fig:teaser_offline}, we evaluate two comparable offline~\cite{mask2former4vis} and online~\cite{huang2022minvis} VIS approaches, both based on the ~\cite{cheng2021mask2former} architecture.
In contrast to IDOL, we do not perform oracle experiments but first split the OVIS training sequences into a train and validation set and then split each validation sequence into sub-sequences of length $\mathcal{T}$.
While the original version of Mask2Former for VIS~\cite{mask2former4vis} is only trained on two frames, we train additional versions on 4 and 8 frames.
%
The superiority of the online MinVIS approach for longer sequences agrees with the findings of IDOL.
However, for sequences similar in length to the number of offline training frames this gap closes, \eg, $\mathcal{T}=8$ and Mask2Former trained on 8 frames.
Furthermore, our results show that training on more frames improves performance on longer sequences.
The experiments of IDOL did not consider the generalization gap between the number of training and test frames.
%
More specifically, their offline candidates~\cite{seqformer, IFC} are trained on 3 and 5 frames, respectively, but expected to generalize to evaluations with up to 36 frames.
Therefore, instead of renouncing offline methods entirely, we advocate for models that predict multiple frames at once but avoid the frame generalization gap via near-online processing.
In~\cref{fig:teaser_near_online}, we show comparable near-online versions.
Their saturation/drop for increasing clip size coincides with observations of previous near-online~\cite{stem_seg, mask_prop} approaches.
However, for longer sequences ($\mathcal{T} > 4$) all presented near-online versions perform as well or better than the online counterpart.



%
%
To support our argument, we present the transformer-based \emph{Near-Online Video Instance Segmentation} (\textbf{\method{}}) method.
Our approach combines offline and online aspects by performing \emph{spatio-temporal instance segmentation} and tracking with \emph{overlap embeddings} jointly in an end-to-end trainable model.
%
%
Given a clip of frames, our model directly predicts spatio-temporal mask volumes and their corresponding object class.
To this end, we compute cross-attention between instance queries and the entire three dimensional feature volume.
%
The model design allows us to bridge short-term object occlusions within a mask volume.
%
To avoid expensive and handcrafted tracking heuristics, we match instances in adjacent clips via cosine similarity of query embeddings~\cite{huang2022minvis,IDOL}.
%
%
%
\cref{fig:teaser_novis} illustrates the relation between clip length and stride/overlap.
%
For long clips with small overlap, adjacent instance queries will not match if an object trajectory changes significantly.
To avoid large and computationally expensive overlaps, we compute overlap embeddings via cross-attention between each instance query and the features of the clip overlap.
%
%

~\method{} is the first end-to-end near-online method that relies not on explicit, handcrafted tracking heuristics like~\cite{IFC, devis} and outperforms comparable online (MinVIS) and offline (IDOL, VITA) 
approaches by a large margin.
We surpass previous methods by \textbf{+3.0} and \textbf{+3.7} points on the \textbf{YouTube-VIS 2019/2021} benchmarks.
For the challenging \textbf{OVIS} dataset, we obtain a new state-of-the-art with an improvement of \textbf{+2.5} points.

\section{Related work}

We differentiate Video Instance Segmentation (VIS) approaches by their respective frame processing paradigms:


\noindent \textbf{Offline.}
Since its inception, the VIS community focused on offline approaches which promise superior track consistency and mask quality over time by processing a given sequence all at once.
The SeqMask-RCNN~\cite{prop_reduce} generates sequence-wide instance proposals based on multiple key frames.
A post-processing removes redundant proposals via intersection over union (IoU).
The authors of~\cite{stmask} improve on single-frame processing via a temporal fusion of adjacent frames.
Initiated by VisTR~\cite{vistr}, the community has shifted towards transformer-based methods.
VisTR extends the DETR~\cite{DETR} object detector to the temporal domain computing attention between all pixels in a spatio-temporal feature volume.
The IFC~\cite{IFC} method reduces expensive attention computations via inter-frame tokens and~\cite{seqformer, heo2022vita, mask2former4vis} share temporal information on the object/instance query level.
In particular, VITA~\cite{heo2022vita} decouples the per-frame segmentation and instance association entirely which allows them to run very long sequences.
The TeViT~\cite{yang2022tevit} approach benefits from an early fusion of temporal information via messenger shift 
tokens in the backbone.

Except from~\cite{mask2former4vis}, existing offline methods share temporal information across multiple frames but refrain from treating VIS as a single spatio-temporal mask prediction problem.
For example, VisTR computes spatio-temporal attention but relies on a single image mask prediction head.
%
%
%
Furthermore, the high memory costs limit the number of training frames which results in \emph{diminishing test-time performance for increasing sequence length}.
Running inference on very \emph{long sequences is infeasible} for most offline methods.
%
%
%
Our near-online model incorporates an end-to-end trainable temporal information exchange and directly predicts spatio-temporal VIS masks without suffering from the aforementioned memory limitations.

\noindent \textbf{Online.}
A frame-to-frame online processing is essential for many real-time applications.
%
During evaluation, these methods must generate tracks on the current frame without including any future information.
%
Early VIS methods~\cite{Yang2019vis, sip_mask, sg_net} generate and connect per-frame mask predictions with a tracking head.
The additional head computes instance embeddings and matches them to a memory/buffer of previous embeddings.
The authors of~\cite{cross_vis} and~\cite{CompFeat} improve on the embedding matching with a crossover learning scheme and feature aggregation via attention, respectively.
The VISOLO~\cite{VISOLO} method propagates information stored in a track memory via grid-based space-time aggregation.
Generally, online methods are expected to be inferior to an offline processing of the entire sequence at once.
However, recently IDOL~\cite{IDOL} and MinVIS~\cite{huang2022minvis} surpassed all previous online and offline methods by segmenting and tracking objects via Transformer instance queries.
IDOL further improves query matching with a contrastive loss between frames.

It is common for online methods to rely on a \emph{heuristic track memory/buffer}.
These buffers recover tracks from shot-term occlusions and reduce the overall noise between frames.
We interpret our near-online processing with clip overlap as an end-to-end trainable version of such a track memory.

\noindent \textbf{Near-online.}
Near-online methods process clips of multiple frames in a sliding window fashion and connect clips via instance tracking.
%
%
%
A significant overlap between clips allows for a more consistent tracking generally superior to online methods.
It is common for online methods to track instances via embedding distance.
Existing near-online approaches, on the other hand, rely on handcrafted heuristics, \eg, mask IoU or object class.
Adding such an instance tracking enables near-online processing with any offline model, see OVIS benchmark results of~\cite{IFC,seqformer, mask2former4vis,yang2022tevit}.
\emph{However, the VIS community is missing specifically designed end-to-end trainable near-online approaches.}
In other words, existing near-online methods merely combine offline processing with a heuristic instance tracking.
%
%
MaskProp~\cite{mask_prop} extends~\cite{Yang2019vis} with a clip-wide mask propagation branch.
The authors of STEm-Seg~\cite{stem_seg} predict pixel embeddings with Gaussian variances to model object instances in spatio-temporal volumes.
The DeVIS~\cite{devis} approach presents an efficient version of VisTR by computing deformable attention. 

We deem our work as the first near-online VIS method which avoids handcrafted heuristics by performing instance tracking via end-to-end trainable embeddings.
To this end, we compute overlap embeddings via cross attention between instance queries and clip overlaps.
%

\section{Problem statement}
\label{sec:problem_statement}

\begin{figure*}[t]
    \centering
    \includegraphics[width=\textwidth]{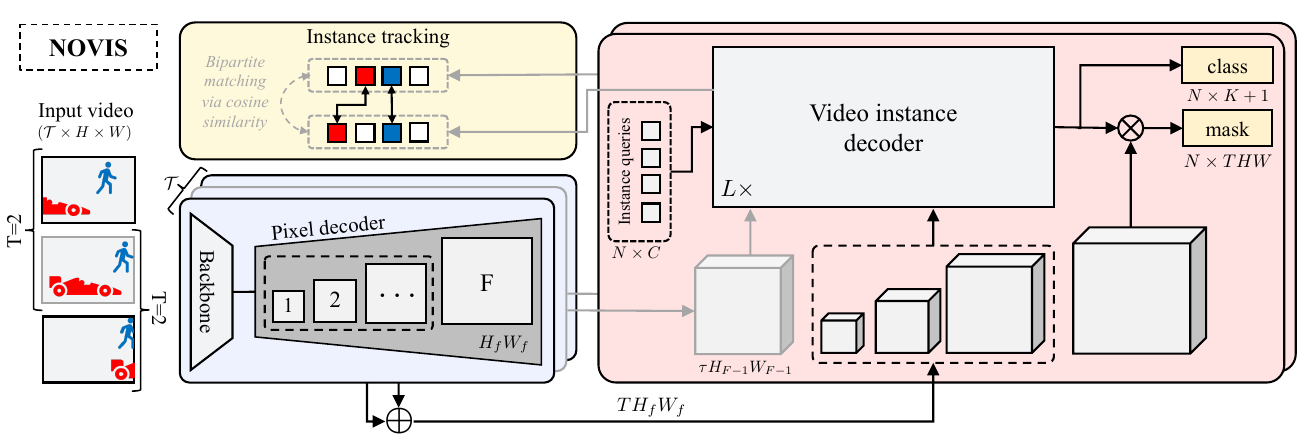}
    \vspace{-6mm}
    \caption{
    Overview of our near-online \textbf{\method{}} model architecture and clip processing pipeline.
    Given a video with length $\mathcal{T}$ and clips with overlap $\tau = [t+S, \dots, t+T]$, our model first computes $F$ per-frame feature scales via deformable attention in the \emph{pixel decoder}.
    The feature scales of a clip are concatenated ($\boldsymbol{\oplus}$) to spatio-temporal volumes ($T H_f W_f$).
    Our \emph{video instance decoder} computes cross-attention between instance queries and feature volumes over $L$ decoding layer.
    The resulting output embeddings directly compute the object class and segmentation mask via a linear layer and dot product ($\boldsymbol{\otimes}$) with the $F$-th feature scale, respectively.
    We track instances between two clips by matching output embeddings via cosine similarity.
    To improve results for objects with large trajectory changes, we match overlap embeddings computed on the shared frame features ($\tau H_{F-1} W_{F-1}$).
    }
    \label{fig:method}
\end{figure*}

%
Given a video sequence $\mathbf{X} \in \mathbb{R}^{\mathcal{T} \times H \times W \times 3}$ of arbitrary length $\mathcal{T}$ consisting of RGB color frames, the Video Instance Segmentation (VIS) task requires the segmentation and temporal consistent tracking of object instances.
In contrast to VOS~\cite{ytvos}, the number of object instances within a sequence is unknown.
A single instance prediction $i$ is defined by its object class $c_i \in \{1, \dots, K\}$ and spatio-temporal foreground-background mask $\mathbf{Y}_i \in  \{0,1\}^{\mathcal{T} \times H \times W}$.
The VIS task does not evaluate any background/stuff classes.
The instance mask $\mathbf{Y}_i$ includes occlusions and measures object identity preservation via Intersection over Union (IoU) to other instance masks.

\section{\method}
\label{sec:method}

%
%

To process a given video sequence most VIS methods either follow an online or offline approach.
Online methods predict object instances one frame at a time and are hence prone to suffer from temporally inconsistent mask and object identities.
Offline methods mitigate these issues by processing the entire video at once but the associated computational costs limit their applicability to long sequences.

Near-online methods, on the other hand, avoid the respective downsides by processing a sequence in clips of frames.
%
To demonstrate their superiority, we present \emph{Near-Online Video Instance Segmentation} (\method{}), an end-to-end trainable near-online model free from any handcrafted tracking heuristics.
%
%
%
This is achieved by combining offline and online aspects into two processing steps: (i) spatio-temporal instance segmentation, and (ii) instance tracking between clips with overlap embeddings.

\subsection{Spatio-temporal instance segmentation}

Given an input clip of size $T \leq \mathcal{T}$, our model directly predicts spatio-temporal instance masks and their object class.
As previous VIS methods~\cite{huang2022minvis,heo2022vita,mask2former4vis}, we build upon the all-purpose mask prediction model introduced in~\cite{cheng2021mask2former}.
Following the recent trend of image detection and instance segmentation approaches,~\cite{cheng2021mask2former} segments objects via Transformer~\cite{attention_is_all_you_need} self- and cross-attention between queries and image features.

Similar to~\cite{mask2former4vis}, we extend this concept to the temporal domain by segmenting three-dimensional object tubes within a clip $\mathbf{X}_{t:t+T} \in \mathbb{R}^{T \times H \times W \times 3}$ of adjacent input frames.
The backbone and pixel decoder, illustrated in~\cref{fig:method}, extract object features per frame and on multiple feature scales.
To this end, the pixel decoder computes deformable attention~\cite{deformable_detr} between pixels and across $F$ feature scales, \eg, $1/32$, $1/16$, $1/8$ and $1/4$ for a ResNet-50~\cite{resnet} backbone.
Given the extracted features, it is the responsibility of the \emph{video instance decoder} to segment individual object trajectories within the clip.
%

\noindent \textbf{Video instance decoder.}
Most existing Transformer-based offline VIS approaches~\cite{seqformer,vistr,IFC,heo2022vita} segment objects over time by temporally aligning sets of \emph{per-frame} instance queries.
In \method{}, we advocate for a more holistic temporal approach by directly predicting three dimensional object masks with a single set of \emph{video instance queries}.
To this end, we compute cross-attention between a set of $N$ queries $\mathbf{Q} \in \mathbb{R}^{N \times C}$ with hidden dimension $C$ and entire spatio-temporal feature volumes with a (flattened) dimension of $TH_{f}W_{f}$.
%
%
More specifically, our instance decoder consists of $L$ decoding layers alternating between instance query self-attention and cross-attention with multi-frame features.
In addition to the common spatial encoding, we add a temporal encoding to each feature volume.
This allows queries to discriminate between frames and build temporally consistent three-dimensional object tubes.
By iterating over the $F$ feature scales, a single instance query computes cross-attention $L / F$ times with each scale.
Following~\cite{cheng2021mask2former}, we apply masked cross-attention based on the predictions of each previous layer.
The spatio-temporal attention for layer $l$ is masked via the re-scaled binary mask prediction $\mathbf{M}_{l-1} \in \{0,1\}^{N \times TH_{l-1}W_{l-1}}$ of the previous ($l-1$)-th instance decoder layer.
Ignoring the zero entries of $\mathbf{M}_{l-1}$ alleviates the foreground-background pixel imbalance allowing the model to focus on the object instances.

The resulting instance output embeddings $\mathbf{Q}_L \in \mathbb{R}^{N \times C}$ directly predict the necessary VIS task outputs $c_i$ and $\mathbf{Y}_i$ without relying on any additional post-processing.
The class predictions $c_i$ are obtained by projecting each embedding into $K + 1$ dimensional class space.
The additional background class allows the model to predict an unknown number of objects within a clip.
We obtain the foreground-background mask logits by computing the dot-product between $\mathbf{Q}_L$ and the $F$-th feature volume.
%
%
%
%
%
%
%
For more details on the underlying segmentation architecture, we refer to~\cite{cheng2021mask2former}.

\noindent \textbf{Class and multi-frame segmentation losses.}
%
%
The loss for a given clip of frames consists of a cross-entropy class $\mathcal{L}_{class}$ and two segmentation losses, namely, a binary mask cross-entropy $\mathcal{L}_{mask}$ and dice $\mathcal{L}_{dice}$ loss.
To compute each loss, we match our model predictions to the ground truth objects present in (at least one frame of) the given clip.
%
%
Previous methods~\cite{mask2former4vis,seqformer,IFC,vistr,heo2022vita} compute both segmentation losses per-frame. 
We found a volumetric version of $\mathcal{L}_{dice}$, which computes a single loss for the entire clip, to be superior.
However, a volumetric version of the per-pixel $\mathcal{L}_{mask}$ loss suffers from a foreground-background pixel imbalance for objects with several occluded frames.
Hence, we compute the mask loss separately for each frame.

In contrast to MinVIS and VITA, our multi-frame segmentation loss includes fully occluded frames.
This allows our model to predict object occlusions as missing part of the three dimensional mask and perform short-term re-identification for windows smaller than $T$.
In other words, our model performs like an end-to-end trainable track buffer/memory commonly applied by online methods~\cite{Yang2019vis,sip_mask,sg_net,IDOL,VISOLO}.

%


\subsection{Instance tracking with embeddings}

The number of training and inference frames within a clip is limited by the computational resources.
As discussed in~\cref{fig:teaser} for offline methods, clip models yield suboptimal performance for sequences significantly longer ($\mathcal{T} \gg T $) than the number of frames it was trained on.
This generalization issue is comparable to training and testing on vastly different image input resolutions.

Hence, we advocate for a near-online instance tracking in a sliding window fashion with stride $1 \leq S \leq T $.
For $S = 1$ the method operates in a pure online fashion.
Averaging mask predictions for overlapping frames further boosts the temporal mask consistency and overall quality.
Finding the optimal clip length $T$ to stride $S$ ratio requires a balancing of two near-online processing effects.
On the one hand, a smaller stride, \ie, larger overlap between two adjacent clips, improves and stabilizes instance tracking and mask quality.
On the other, a near-online processing with minimum overlap larger than zero ($S = T - 1$) increases the potential re-identification window between two clips up to $2T - 1$.
In our experiments, we will show ablations on optimal clip length to stride ratio.

%
%
%
%

Existing near-online VIS approaches rely on handcrafted tracking heuristics, \eg, mask IoU or object class, to connect instances between clips.
However, for online VIS approaches~\cite{huang2022minvis,IDOL}, a cosine distance matching between decoder queries of adjacent frames has shown superior results to the common heuristics.
Our video instance decoder allows us to apply a similar matching between instance queries of two adjacent clips.
However, for objects with large visual or trajectory changes and clips with small overlap the risk of matching wrong instance queries increases.
~\cref{fig:clip_length_and_stride_analysis} shows that in particular long clips, \eg, for $T=6$ on OVIS~\cite{ovis}, suffer from this effect.
%
%
To minimize the necessary overlap and maximize the benefits from potential shot-term re-identifications between two clips, we compute \emph{overlap embeddings} specifically designed for a near-online instance tracking.


\subsection{Clip overlap embeddings}
%
In~\cref{fig:method}, we illustrate the integration of overlap embeddings into the overall model architecture.
%
%
At its core, the clip-to-clip tracking is performed via cosine similatity but we do not directly match output embeddings $\mathbf{Q}^{t}_L= \mathbf{Q}^{t:t+T}_L \in \mathbb{R}^{N \times C}$ and $\mathbf{Q}^{t+S}_L = \mathbf{Q}^{t+S:t+S+T}_L \in \mathbb{R}^{N \times C}$ of two adjacent clips with length $T$.
Instead, we compute the representation of each embedding with respect to the overlapping frames $\tau = [t+S, \dots, t+T]$ at the start and end of their respective clip and perform inter-clip matching only based on these shared frames.
Performing near-online instance tracking only based on overlapping frames is a common practice for heuristic approaches and often achieved by computing a volumetric IoU~\cite{mask_prop,stem_seg,devis}.

To obtain the instance representation $\mathbf{Q}^{t \rightarrow \tau}_L$ of the subset of frames $\tau$ of the clip starting at $t$, we compute cross-attention between each output embedding and the respective frame features.
%
%
%
More specifically, we re-compute the last instance decoder layer $L$ with the following attention mask for the $n$-th query:
\begin{align}
\mathbf{\hat{M}}_{L} (n, t, x, y) = \left\{\begin{array}{ll}
  \mathbf{M}_{L}(n, t, x, y)  & \text{if~} t \in \tau \\
   0 & \text{otherwise}
\end{array}\right..
\end{align}
If a query predicts no masks within the overlap, \eg, due to an object occlusion, the equation above results in $\mathbf{\hat{M}}_{L} (n, t) = 0~\forall~t $.
In that case, we refrain from computing the overlap attention and use the original output embedding for an instance matching based on the entire clip.

The video tracking module allows for a computation of instance representations on the edges of each clip more suitable for embedding-based matching.
Exploiting the clip overlap allows us to avoid additional contrastive learning between embeddings as done in IDOL or VITA.

\begin{table*}[t]
\centering
\resizebox{\textwidth}{!}{
\begin{tabular}{ll|c|c|r|ccccc|ccccc HHH}
\toprule
\multicolumn{2}{c|}{\multirow{2}*{Method}} & \multicolumn{1}{c|}{\multirow{2}*{Data}} & \multicolumn{1}{c|}{\multirow{2}*{$T$ / $S$}} & \multicolumn{6}{c}{YouTube-VIS 2019~\cite{Yang2019vis}} & \multicolumn{5}{|c}{YouTube-VIS 2021~\cite{Yang2019vis}} & \multicolumn{3}{H}{OVIS~\cite{ovis}}\\
\cmidrule{5-18}
& & & & \multicolumn{1}{c|}{FPS} & AP & AP$_{50}$ & AP$_{75}$ & AR$_1$ & AR$_{10}$ & AP & AP$_{50}$ & AP$_{75}$ & AR$_1$ & AR$_{10}$ & AP & AP$_{50}$ & AP$_{75}$\\

\midrule

\multirow{15}*{\rotatebox[origin=c]{90}{ResNet-50}}

& MaskTrack~\cite{Yang2019vis}         & VIS       & ON  & 26.1 & 30.3 & 51.1 & 32.6 & 31.0 & 35.5 &  28.6 & 48.9 & 29.6 & 26.5 & 33.8 & $15.4$ & $33.9$ & $13.1$ \\
& VisTR~\cite{vistr}                        & VIS       & OFF & 69.9  & 36.2 & 59.8 & 36.9 & 37.2 & 42.4 & -- & -- & -- & -- & -- & -- & -- & --\\
& IFC~\cite{IFC}                            & VIS & 5 / 1 & 46.5 & 39.0 &60.4 &42.7 & 41.7 & 51.6 & -- & -- & -- & -- & -- & -- & -- & --\\

& IFC~\cite{IFC}                            & VIS       & OFF & 107.1 & 41.2 & 65.1 & 44.6 & 42.3 & 49.6 & 35.2 & 57.2 & 37.5 &-- &-- & -- & -- & --\\

& DeVIS~\cite{devis}                            & VIS       & 6 / 4 & 18.4 & 44.4 & 66.7 & 48.6 & 42.4 & 51.6 & 43.1 & 66.8 & 46.6 & 38.0  & 50.1 & -- & -- & --\\

& SeqFormer~\cite{seqformer}                & VIS       & OFF & 72.3 & 45.1 & 66.9 & 50.5 & 45.6 & 54.6 & -- & -- & -- & -- & -- & -- & --\\
& IDOL~\cite{IDOL}                          & VIS       & ON  & 30.6 & 46.4 & 70.7 & 51.9 & 44.8 & 54.9 & 43.9 & 68.0 & 49.6 & 38.0 & 50.9 & X & X & X \\
& Mask2Former~\cite{mask2former4vis}        & VIS       & OFF &-- & 46.4 & 68.0 & 50.5 &-- & -- & 40.6  & 60.9 &41.8 &-- & -- &-- &-- & --\\

& TeViT~\cite{yang2022tevit}                & VIS       & OFF &-- & 46.6 & 71.3 & 51.6 & 44.9 & 54.3 & 37.9  & 61.2 & 42.1 & 35.1 & 44.6 & -- & -- & --\\

& SeqFormer~\cite{seqformer}                & CC+VIS    & OFF & 72.3 & 47.4 & 69.8 & 51.8 & 45.5 & 54.8 & 40.5 & 62.4 &43.7 &36.1 & 48.1 & -- & --\\
& MinVIS~\cite{huang2022minvis}             & VIS       & ON  & - & 47.4 & 69.0 & 52.1 & 45.7 & 55.7 & 44.2 & 66.0 & 48.1 & 39.2 & 51.7 & 25.0 & 45.5 & 24.0   \\
& IDOL~\cite{IDOL}                                      & CC+VIS    & ON  & 30.6 & 49.5 & 74.0 & 52.9 & 47.7 & 58.7 & -- & -- & -- & -- & -- & -- & -- & -- \\
& VITA~\cite{heo2022vita}                                  & CC+VIS    & OFF & -- & 49.8 & 72.6 & 54.5 & 49.4 & \textbf{61.0} & 45.7 & 67.4 & 49.5 & 40.9 & 53.6 & -- & -- & -- \\

\cmidrule{2-15}

%

& \multirow{2}*{\textbf{\method{}}}                  & VIS & 4 / 2 & 19.9 & \avercalc[1]{52.27, 51.28, 50.78} & \avercalc[1]{75.76, 74.45, 74.61} & \avercalc[1]{56.63, 56.06, 52.91} & \avercalc[1]{49.67, 48.19, 48.15} & \avercalc[1]{60.8, 59.82, 59.58} & \textbf{\avercalc[1]{47.6, 47.4, 46.6}} & \textbf{\avercalc[1]{69.7, 69.9, 68.7}} & \textbf{\avercalc[1]{50.6, 49.6, 49.8}.0} & \textbf{\avercalc[1]{41.3, 41.2, 41.5}} & \textbf{\avercalc[1]{54.6, 54.9, 53.8}} & X & X & X \\

&                  & CC+VIS & 4 / 2 & 19.9 & \textbf{52.8} & \textbf{75.7} & \textbf{56.9} & \textbf{50.3} & 60.6 & -- & -- & -- & -- & -- & -- & -- & -- \\

\midrule
\multirow{8}*{\rotatebox[origin=c]{90}{Swin-L}}

& SeqFormer~\cite{seqformer}                            & VIS    & OFF & 27.7 & -- & -- & -- & -- & -- & 51.8 & 74.6 & 58.2 & 42.8 & 58.1 & -- & -- & -- \\

& DeVIS~\cite{devis}                            & VIS       & 6 / 4 & -- & 57.1 & 80.8 & 66.3 & 50.8 & 61.0 & 54.4 & 77.7 & 59.8 & 43.8  & 57.8 & -- & -- & --\\

& SeqFormer~\cite{seqformer}           & CC+VIS    & OFF & 27.7 & 59.3 & 82.1 & 66.4 & 51.7 & 64.4 & -- & -- & -- & -- & -- & -- & -- & -- \\

& IDOL~\cite{IDOL}                      & VIS       & ON & 17.6 & 61.5 & 84.2 & 69.3 & 53.3 & 65.6 & 56.1 & 80.8 & 63.5 & 45.0 & 60.1 & X & X & X \\
& VITA~\cite{heo2022vita}                                  & CC+VIS    & OFF & -- & 63.0 & 86.9 & 67.9 & 56.3 & 68.1 & -- & -- & -- & -- & -- & -- & -- & -- \\
& IDOL~\cite{IDOL}                                  & CC+VIS    & ON & 17.6 & 64.3 & 87.5 & 71.0 & 55.6 & 69.1 & -- & -- & -- & -- & -- & -- & -- & -- \\

\cmidrule{2-15}

& \multirow{2}*{\textbf{\method{}}}                  & VIS & 4 / 2 & 10.0 & \avercalc[1]{64.29, 63.35, 63.84} & \avercalc[1]{86.44, 87.15, 86.29} & \avercalc[1]{71.38, 68.79, 67.66} & \avercalc[1]{55.87, 56.08, 55.62} & \avercalc[1]{69.11, 68.66, 68.22} & \textbf{\avercalc[1]{59.7, 59.8, 59.8}} & \textbf{\avercalc[1]{81.5, 82.1, 82.3}.0} & \textbf{\avercalc[1]{66.7, 66.9, 65.8}} & \textbf{\avercalc[1]{47.7, 48.0, 47.9}} & \textbf{\avercalc[1]{63.8, 65.0, 64.4}} & X & X & X \\

&                                                    & CC+VIS & 4 / 2 & 10.0 & \textbf{65.7} & \textbf{87.8} & \textbf{72.2} & \textbf{56.3} & \textbf{70.3} & -- & -- & -- & -- & -- & -- & -- & -- \\

\bottomrule
\end{tabular}
}

\caption{
Benchmark results on the~\textbf{YouTube-VIS 2019 and 2021} validation sets.
We sort results by the YouTube-VIS 2019 average precision (AP) and distinguish methods between online (ON), offline (OFF) and near-online (clip length $T$ and stride $S$).
Additional training data by simulating clips via COCO image augmentations is denoted with \textit{CC+VIS}.
%
%
%
%
%
}

\label{tab:eval_ytvis_19_21_ovis}

\end{table*}

\section{Experiments}
\label{sec:experiments}

\subsection{Benchmarks}

\noindent \textbf{YouTube-VIS 2019 and 2021.}
The YouTube-VIS~\cite{Yang2019vis} datasets provide instance ground truth masks for 40 object categories on a total of 2883 and 3859 sequences, respectively.
Although less severe for the 2021 version, the YouTube-VIS benchmark only includes comparatively short videos (average of 27.4 / 39.7 frames for the validation sets).
Furthermore, both versions provide comparatively few object instances (average of 1.69 / 2.10 per video) and include little to no object occlusions and movement.
In particular the short sequences allowed offline approaches to flourish and dominate the VIS community.

\noindent \textbf{OVIS.}
The Occluded VIS (OVIS)~\cite{ovis} benchmark follows the same data structure as~\cite{Yang2019vis} but includes significantly longer sequences (average of 62.7 for the validation set) crowded with more object instances (average of 5.80 per video).
The benchmarks labels 25 object categories and currently represents the most challenging public VIS benchmark.
To test the limits of our ~\method{} approach and near-online processing in general, we perform ablations on~\cite{ovis}.

\noindent \textbf{Metrics.}
Following~\cite{Yang2019vis}, all VIS benchmarks evaluate average precision (AP) and recall (AR).
The ground truth matching is performed via volumetric IoU matching with varying thresholds (0.5 and 0.75).

\subsection{Implementation and training details}
Our~\method{} method builds on top of the Mask2Former~\cite{cheng2021mask2former} image segmentation model.
If not otherwise specified, we follow their hyperparameter settings.
Each model is pre-trained on COCO~\cite{COCO} as in~\cite{cheng2021mask2former} and then fine-tuned on the respective VIS dataset.
For YouTube-VIS 2019 it is common to additionally report results from a joint COCO and VIS (CC+VIS) training.
To this end, we follow the augmentation pipeline of~\cite{heo2022vita} to simulate clips from COCO images.
%
%
%
We train with a batch size of 8 distributed over 8 $\times$ 32 GB GPUs and apply random clip frame reversal, horizontal flipping, cropping and rescaling (shortest edge between 228 and 512 pixels) augmentations.
Following the VIS convention, the minimum shortest edge size at test-time (MST) is 360 pixels.
For OVIS, we also report results augmented from 480 to 800 and evaluated with 720 pixels.
%
%
We ablate with a ResNet-50~\cite{resnet} backbone on a 2-fold OVIS training set split.
All reported results are averaged over three runs with different random seeds.
For more details and benchmark specific hyperparameters, we refer to the supplementary.

\subsection{Benchmark results}

\begin{table}[t]

\resizebox{\columnwidth}{!}{
\centering
\begin{tabular}{ll|Hc|c|Hccccc}
\toprule
\multicolumn{2}{c|}{Method} & Data & $T$ / $S$ & MST & Top-K & AP & AP$_{50}$ & AP$_{75}$ & AR$_{1}$ & AR$_{10}$ \\

\midrule

\multirow{11}*{\rotatebox[origin=c]{90}{ResNet-50}}

& MaskTrack~\cite{Yang2019vis}    & VIS        & ON & -   & -                        & \hfill 10.8 & 25.3 & 8.5  & 7.9  & 14.9 \\
& IFC~\cite{seqformer}                 & VIS        & 10 / 1    & -   & -          & \hfill 13.1 & 27.8 & 11.6 & 9.4  & 23.9 \\
& SeqFormer~\cite{seqformer}           & VIS        & 10 / 1    & 720   & -        & \hfill 15.1 & 31.9 & 13.8 & 10.4 & 27.1 \\
& Mask2Former~\cite{mask2former4vis}   & VIS        & 30 / 30  & 360   & 10        & \hfill 17.3 & 37.3 & 15.1 & 10.5 & 23.5 \\
& TeViT~\cite{yang2022tevit}  & VIS        & 36 / -   & 360   & -         & \hfill 17.4 & 34.9 & 15.0 & 11.2 & 21.8  \\ 

& VITA~\cite{heo2022vita}  $^{\ast\ast}$            & CC + VIS   & OFF      & 360   & 20        & \hfill 19.6 & 41.2 & 17.4 & 11.7 & 26.0 \\

& DeVIS~\cite{devis}                            & VIS       & 6 / 4 & 360   & 30        & \hfill 23.8 & 48.0 & 20.8 & -- & -- \\

& MinVIS~\cite{huang2022minvis}        & VIS        & ON       & 360   & 10        & \hfill 25.0 & 45.5 & 24.0 & 13.9 & 29.7 \\
& IDOL~\cite{IDOL}                     & VIS        & ON       & 720   & thresh    & \hfill 30.2 & 51.3 & 30.0 & 15.0 & \textbf{37.5} \\

\cmidrule{2-11}


& \multirow{2}*{\textbf{\method{}}}                   & VIS          & 4 / 2 & 360 & 30 & \hfill 30.8 & 54.4 & 31.0 & 15.2 & 35.3 \\

&                                                     & VIS          & 4 / 2 & 720 & 30 & \hfill \textbf{32.7} & \textbf{56.2} & \textbf{32.6} & \textbf{15.7} & 37.1 \\


\midrule
\multirow{6}*{\rotatebox[origin=c]{90}{Swin-L}} 

& Mask2Former~\cite{mask2former4vis}   & VIS        & 30 / 30  & 360   & 10        & \hfill 25.8 & 46.5 & 24.4 & 13.7 & 32.2 \\

& VITA~\cite{heo2022vita} $^{\ast\ast}$             & CC + VIS   & OFF      & 360   & 20        & \hfill 27.7 & 51.9 & 24.9 & 14.9 & 33.0 \\

& DeVIS~\cite{devis}                            & VIS       & 6 / 4 & 360   & 30        & \hfill 34.6 & 58.7 & 36.8 & -- & -- \\

& MinVIS~\cite{huang2022minvis}        & VIS        & ON       & 360   & 10        & \hfill 39.4 & 61.5 & 41.3 & 18.1 & 43.3 \\
& IDOL~\cite{IDOL}                     & VIS        & ON       & 720   & thresh    & \hfill 42.6 & 65.7 & 45.2 & 17.9 & \textbf{49.6} \\

\cmidrule{2-11}

& \multirow{2}*{\textbf{\method{}}}                  & VIS & 4 / 2 & 360 & 30 & \hfill 43.0 & 66.9 & 44.5 & 18.9 & 46.3 \\
&                                                    & VIS & 4 / 2 & 720 & 30 & $^{\ast}$\textbf{43.5}  & \textbf{68.3} & \textbf{43.8} & \textbf{19.4} & 46.9 \\

\bottomrule
\end{tabular}
}

\caption{
    Benchmark results on the~\textbf{OVIS} validation set.
    We indicate the minimum size at test-time (MST) of the shortest frame edge and distinguish between online (ON), offline (OFF) and near-online (clip length $T$ and stride $S$).
    Our model with $^{\ast}$ was only evaluated but not trained on larger (720) input resolutions.
    With $^{\ast\ast}$ we denote a joint training on simulated clips from COCO.
    Mask2Former and SeqFormer/IFC results are from the MinVis and IDOL papers.
    %
    %
}

\label{tab:eval_ovis}

\end{table}

\noindent \textbf{YouTube-VIS 2019 and 2021.}
The presented~\method{} method achieves state-of-the-art results on both YouTube-VIS~\cite{Yang2019vis} benchmarks surpassing all previous methods by large margins.
In~\cref{tab:eval_ytvis_19_21_ovis}, we present results for our best performing near-online configuration trained only on YouTube-VIS or jointly with COCO (CC+VIS). 
We differentiate between online (ON), offline (OFF) and near-online with clip length $T$ and stride $S$.
We outperform IDOL, the previous best online method, by \textbf{+4.5} and \textbf{+3.7} points on the 2019 and 2021 benchmarks, respectively.
To demonstrate the superiority of near-online processing, we want to focus on the offline (VITA / Mask2Former) and online (MinVIS) methods which also build on top of~\cite{cheng2021mask2former}.
Most notably, our approach outperforms both offline methods (\textbf{+3.0} and \textbf{+4.5} on YouTube-VIS 2019) as they mitigate their computational limitations by either avoiding any temporal pixel-level information exchange (VITA) or training on far less frames (only 2) than required during test-time (Mask2Former).
While offline methods which are able to process all frames of short sequences at once generally obtain higher runtimes, our~\method{} approach still achieves 19.9 FPS.
The superiority of our approach also translates to the top-performance evaluation setting with a Swin-L~\cite{SwinTransformer} backbone.
%

\noindent \textbf{OVIS.}
To further strengthen the argument for near-online approaches, we present state-of-the-art results on the challenging OVIS~\cite{ovis} benchmark in~\cref{tab:eval_ovis}.
For an increased train and test resolution (MST), we outperform the runner up IDOL method by \textbf{+2.5} points.
The significantly longer sequences and increased object occlusions present in OVIS only exacerbate the aforementioned issues of existing offline methods.
Hence, we outperform VITA by \textbf{+11.2} points.
The second strongest near-online approach, DeVIS, suffers from its heuristic instance tracking which, as we will show in our ablations, is substantially inferior to our embedding-based tracking.
However, the increased input resolution of IDOL highlights a weakness that our approach inherits from offline methods.
For a Swin-L backbone and $T$$=$$4$ our method exceeds the computational limits during training.
Hence, we present a version only evaluated on the higher input resolution.


\begin{figure}[t]
    \centering
    \includegraphics[width=0.95\columnwidth]{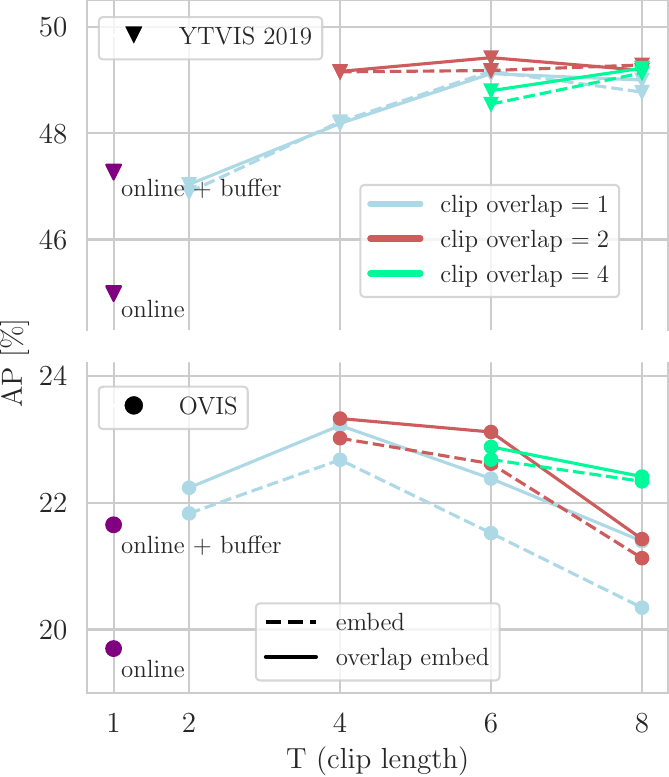}
    \caption{
        Analysis on \textbf{clip length} and \textbf{stride/overlap} evaluated on training set splits of YouTube-VIS 2019 and OVIS.
        Instance tracking between clips is performed via cosine distance matching.
        %
        Computing cross-attention between queries and clip overlaps improves performance for larger clip sizes.
        }
    \label{fig:clip_length_and_stride_analysis}
\end{figure}

\subsection{Ablation studies}
%
%

In order to predict instance segmentations over time, our~\method{} method treats individual frames as a single spatio-temporal mask volume.
Our baseline with $20.9$ AP in~\cref{tab:ablation_novis} represents a naive application of an image segmentation method~\cite{cheng2021mask2former} to such a pixel volume.
The heuristic tracking incorporates multiple cues based on volumetric IoU and object class.
%
%
%
Adding a temporal encoding allows instance queries to discriminate between frames and boosts results by $+0.8$.
For objects occluded in several frames, the volumetric mask loss exacerbates the foreground-background pixel imbalance.
Hence, our final model only computes volumetric versions of the dice loss.
Replacing the handcrafted and expensive heuristics with embedding matching allows for an implicit multi-cue tracking improving results by $+1.1$ points.
%
%
We evaluate results with comparatively large clip overlaps.
Hence, a computation of overlap embeddings yields only $+0.3$ points.
For an in-depth analysis of the relation between clip length, stride and overlap, we refer to the next section.

\begin{table}[t]
    \resizebox{\columnwidth}{!}{
        \centering
        \begin{tabular}{c|cc|c|cccc}
            \toprule
            
            
            & \multicolumn{2}{c|}{volumetric loss} &  \\
            \midrule
            TE & dice & mask & tracking & AP & AP$_{75}$ & AR$_{1}$ \\
            
            \midrule            
                                                &          &          & heuristic          & \avercalc[1]{20.82, 20.8, 20.94} & \avercalc[1]{19.822, 19.8555, 19.4277} & \avercalc[1]{11.9303, 12.2376, 11.8539}.0  \\
                                                
                                                
                              $\times$          &          &          & heuristic          & \avercalc[1]{22.21, 21.78, 21.17} & \avercalc[1]{21.4463, 21.1338, 21.7282} & \avercalc[1]{12.0826, 12.2699, 12.1069} \\
                              $\times$          &          & $\times$ & heuristic          & \avercalc[1]{21.86, 21.37, 21.03} & \avercalc[1]{20.6507, 21.1905, 21.4331} & \avercalc[1]{11.7211, 12.1803, 12.0288}.0 \\
                              
                              $\times$          & $\times$ &           & heuristic         & \avercalc[1]{22.3, 21.1, 22.3} & \avercalc[1]{21.4752, 21.3889, 21.2928} & \avercalc[1]{12.5205, 12.3697, 11.9399} \\
                              
                              $\times$          & $\times$ &           & embed             & \avercalc[1]{23.56, 22.94, 22.55}.0 & \avercalc[1]{22.8299, 23.2890, 23.5173} & \avercalc[1]{12.8319, 12.8159, 12.6353} \\

                              $\times$          & $\times$ &           & overlap embed     & \textbf{\avercalc[1]{24.0378, 23.4938, 22.5071}} & \textbf{\avercalc[1]{23.4318, 23.9994, 22.3686}} & \textbf{\avercalc[1]{13.1931, 12.8372, 12.9671}.0} \\
                              
            
        
            \bottomrule
        \end{tabular}
    }
    \caption{
        Ablation of our near-online \textbf{~\method{}} method on a OVIS training set split.
        We motivate several design decisions from a naive application of Mask2Former without temporal encoding (TE) to our final model evaluated with clip length $T$$=$$4$ and stride $S$$=$$2$.
        %
    }
    \label{tab:ablation_novis}
\end{table}

\subsection{Clip analysis}
To further understand the benefits and limitations of our near-online approach, we evaluate varying clip configurations for OVIS and YouTube-VIS 2019.
The long sequences of OVIS render a direct comparison with offline methods infeasible.
Only~\cite{heo2022vita} is able to process OVIS in a pure offline fashion.
However, their benchmark results in~\cref{tab:eval_ovis} are inferior to MinVis which corresponds to the online variant \emph{without} track buffer in~\cref{fig:clip_length_and_stride_analysis}.

\noindent \textbf{Online with track buffer.}
Adding a track buffer which matches embeddings of the current frame with the average of previous frame embeddings improves the vanilla online version significantly.
This track buffer is comparable to what existing online approaches~\cite{IDOL,VISOLO} apply.
Interestingly, the optimal buffer length for OVIS coincides with our optimal clip length $T=4$.
Nevertheless, our near-online approach outperforms the \emph{online + buffer} version on both benchmarks by large margins.
This is due to the buffer`s inability to improve the object mask quality over time.
Only a (near-online) multi-frame processing provides improved track management \emph{and} more consistent mask predictions without relying on more complex, handcrafted heuristics.

\noindent \textbf{Overlap embeddings.}
For the more challenging OVIS dataset, the cross-attention between instance embeddings and overlapping frames yields substantial improvements.
Large visual or trajectory changes between two clips limit the matchability of their instance queries.
Hence, we see the largest gains of up to 2 points for long clip sizes with small overlap, \eg, $T$$=$$8$ and clip overlap of 1.

\noindent \textbf{Long clip sizes.}
Intuitively, we expect longer clip sizes $T$ to yield better performance as this allows the model to bypass longer occlusions and increase the temporal consistency of object masks.
However, due to the increased object movement and occlusions present in OVIS, we observe degrading performance for clips longer than 4. 
To explain this, one has to understand that for single image segmentation~\cite{cheng2021mask2former} the notion of an object instance emerges from the pixel decoder.
The instance queries merely organize and rank the provided instances.
To predict spatio-temporal instances for challenging object trajectories, the pixel decoder is therefore required to (efficiently) compute temporal connections.
We deem temporal deformable attention as in~\cite{devis,zhou2022transvod} very promising but leave the exploration of further improvements for near-online open for future research.

Nevertheless, for short clip sizes we see significant improvements over the presented online variants.
This indicates that a few frames are already enough to capture the bulk of advantages expected from multi-frame (offline) processing.
We obtain the final near-online configuration ($T$$=$$4$ and $S$$=$$2$) for our benchmark results by evaluating the top configurations of~\cref{fig:clip_length_and_stride_analysis} on the OVIS validation set.







%


%



\section{Conclusion}
We have shown that the recent trend of online methods for VIS must not be understood as an argument against the efficiency of existing offline methods.
In fact, the analysis provided in this work and our state-of-the-art benchmark results make a strong case for multi-frame processing.
Hence, we advocate for near-online methods, which combine the best of online and offline approaches, while avoiding their downsides. 
To this end, we presented \method{} which directly predicts spatio-temporal temporal mask volumes and performs tracking between clips via overlap embeddings.
Our approach represents the first end-to-end trainable near-online method which avoids handcrafted tracking heuristics.
We hope this work fosters future research on dedicated near-online models and marks a valuable rebuttal on the recent online vs. offline debate within the VIS community.



\bibliography{egbib}
\bibliographystyle{icml2023}

\newpage
\appendix
\onecolumn
\section{Implementation and training details}
We only mention the hyperparameters of our~\method{} approach which differ from the default Mask2Former~\cite{cheng2021mask2former} settings.
~\cref{tab:hyperparams_benchmarks} summarizes the total number of training iterations, learning rate drops and top-k evaluation for the three benchmarks presented in this work.
Since~\method{} is designed for near-online processing, its instance decoder is not required to generalize to arbitrary sequence lengths.
This allows us to combine a common sinusoidal spatial with a learned/parametric temporal encoding.
%
%
As mentioned in the main paper, we train our model on normal (228 to 512) and large (480 to 800) input resolution scales.
The former is only evaluated for a comparison to IDOL~\cite{IDOL} on OVIS~\cite{ovis}.
We adjust the number of PointRend~\cite{kirillov2019pointrend} training points for the two scale ranges to 3136 and 6272, respectively.
All Swin-L~\cite{SwinTransformer} models are evaluated with a minimum frame size at test-time (MST) of 480 pixels.
 
To train a model for a specific clip size, we augment the input data by sampling $T$ random frames from a given sequence.
This is achieved by sampling $T-1$ frames (without replacement) in a range of $\pm 5$ frames around a random reference frame.
Considering the OVIS~\cite{ovis} training set contains sequences of up to 500 frames, the sampling range is quite small.
Larger ranges, \ie, more challenging training clips, generally result in worse performance.

\section{Mask quality ablation studies}

\begin{figure}
    \centering
    \includegraphics[width=0.65\columnwidth]{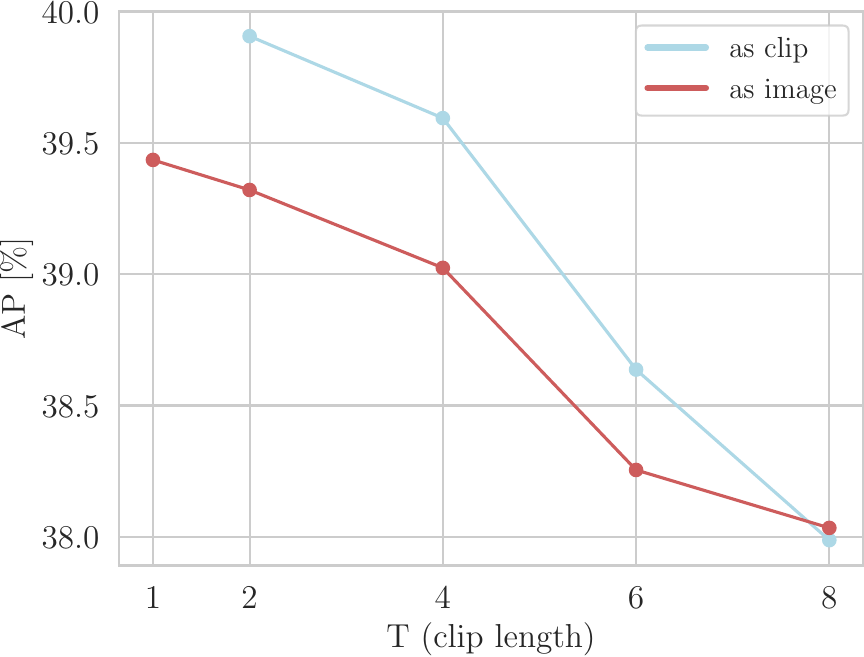}
    \caption{
        Analysis on \textbf{mask prediction quality} for different clip lengths $T$.
        Each model is trained and evaluated on a 2-fold split on the OVIS~\cite{ovis} training set.
        We evaluate per-frame mask prediction performance by processing video data with the same model either as single images or clips of size $T$.}
    \label{fig:clip_length_mask_quality}
\end{figure}

To further demonstrate the superiority of multi-frame processing, \ie, offline and near-online, over frame-by-frame online methods, we evaluate the image segmentation performance of models trained for varying clip sizes in~\cref{fig:clip_length_mask_quality}.
To this end, we process the same video data (2-fold split on the OVIS~\cite{ovis} training set) either as single images or in clips of size $T$ and evaluate the mask average precision for each frame separately.
This allows us to isolate and analyse the mask quality without considering the track consistency or effects of the embedding matching.
Interestingly, we observe diminishing \emph{per-image} segmentation performance with increasing clip size coinciding with our evaluations on full length sequences shown in the main paper.
Training on longer clips (or with a large frame sampling range) has a detrimental effect on the model.
As discussed in the main paper, we consider the inability of the pixel decoder to model large object movements as the main reason for the diminishing performance.


%


%

However, for clip sizes up to $T=6$, we observe a clear advantage of multi-frame processing.
Evaluating a model on clips of frames provides more consistent and less noisy objects masks.
It should be noted, that for $T=4$, our top-performing final benchmark configuration, the image segmentation performance without multi-frame processing is inferior to an (online) model trained for clip size one.
However, for an evaluation on full sequences the advantages of a multi-frame processing outweigh the detrimental effect on the isolated per-image performance.
%
Hence, we see great potential in addressing the diminishing performance of our approach by improving its temporal modelling capacities.




\section{Qualitative benchmark results}
We present qualitative results (including failure cases) for the YouTube-VIS 2019/2021~\cite{Yang2019vis} and OVIS~\cite{ovis} benchmarks.
For visualization purposes, we do not show all top-k results but filter instance predictions with a classification score smaller than 5\%.
Furthermore, the selected examples are chosen with data privacy in mind and do no contain full person objects or visible faces.

\subsection{YouTube-VIS 2019 and 2021}
In~\cref{fig:qualitative_results_ytvis_19_21}, we show results for a range of YouTube-VIS sequences for different object classes. 
Our approach is able to segment and track most object instances and provides detailed masks across multiple frames.
The first 5 rows demonstrates the ability of~\method{} to track various object scales and movements for low (less than 3 object instances) crowdedness.
In particular, row 5 (hand and lizard) shows a successful object-object occlusion handling between different object classes.
Row 7 and 8 illustrate examples for significantly more crowded scenes.
For small object movements, our multi-frame processing with clip overlap is very robust and only rarely produces noisy/inconsistent masks as in the last frame of row 8 (cows).
The last row (bear) demonstrates the limited long-term (more than 60 frames in this example) occlusion handling capabilities of our approach.
However, an additional appearance-based re-identification step or more long-term frame processing could mitigate these issues.

\subsection{OVIS}
\begin{figure*}[t]
    \centering
    
    \includegraphics[width=0.24\textwidth]{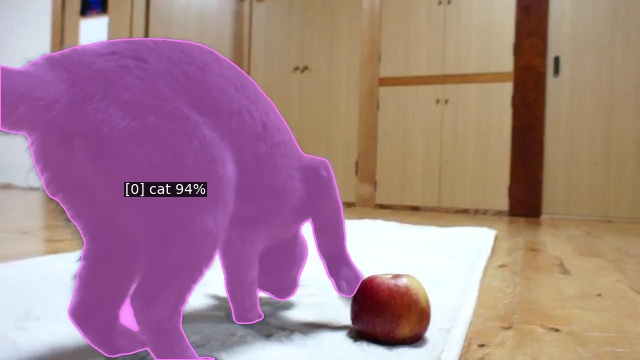}
    \includegraphics[width=0.24\textwidth]{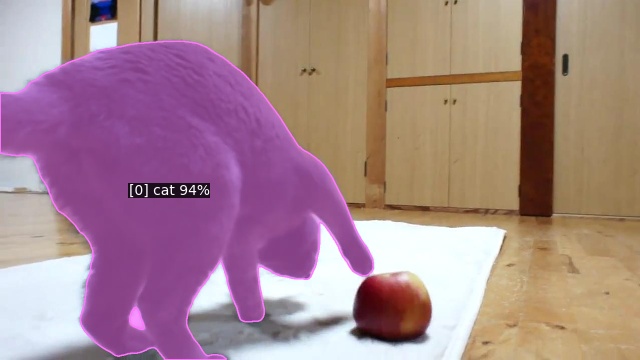}
    \includegraphics[width=0.24\textwidth]{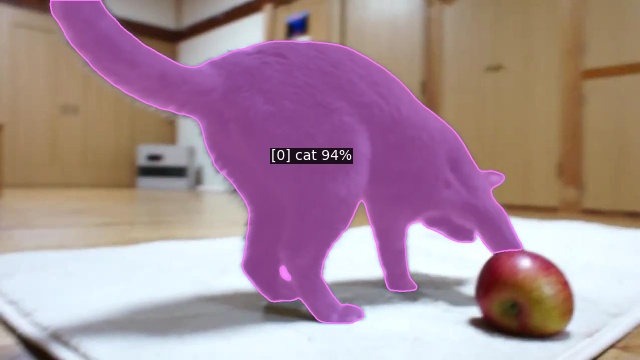}
    \includegraphics[width=0.24\textwidth]{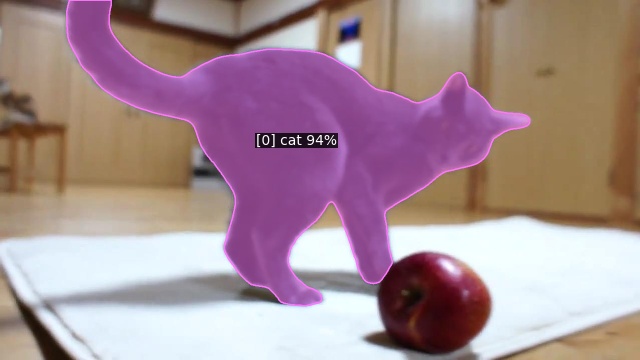}

    \includegraphics[width=0.24\textwidth]{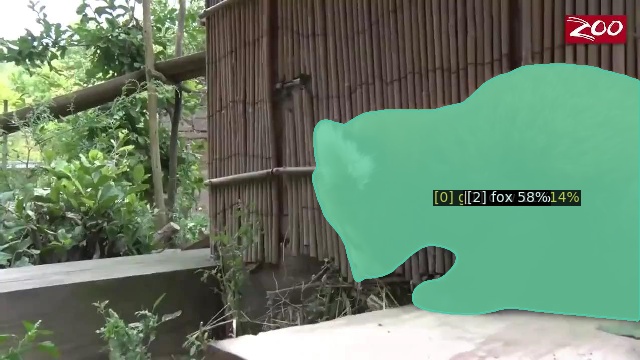}
    \includegraphics[width=0.24\textwidth]{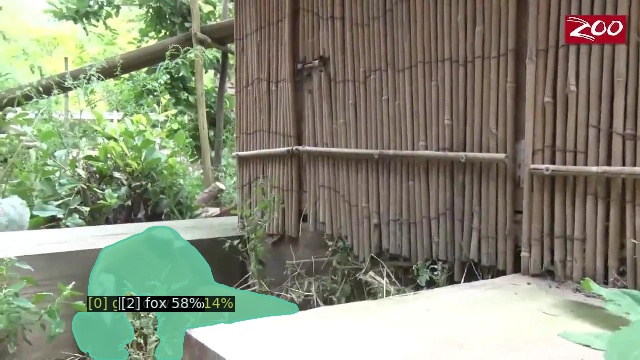}
    \includegraphics[width=0.24\textwidth]{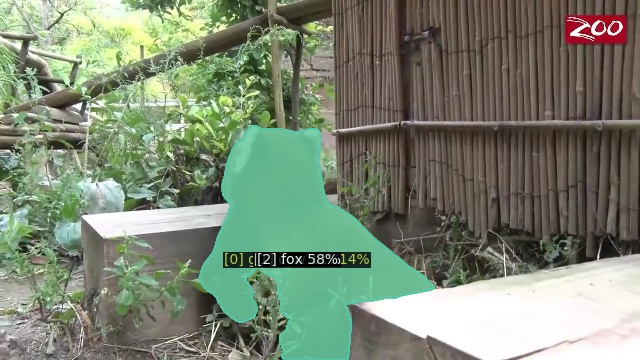}
    \includegraphics[width=0.24\textwidth]{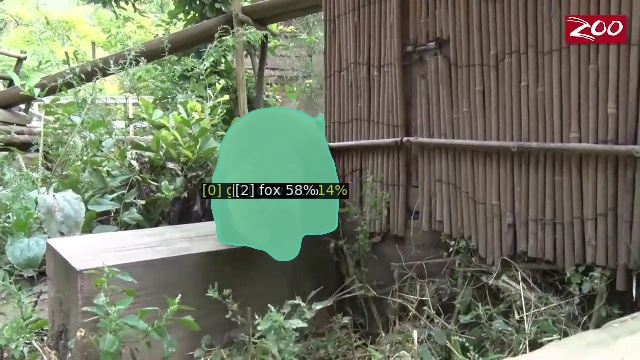}

    \includegraphics[width=0.24\textwidth]{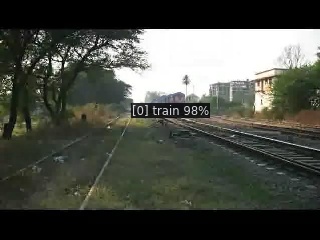}
    \includegraphics[width=0.24\textwidth]{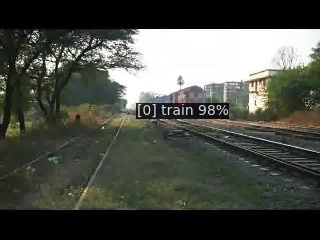}
    \includegraphics[width=0.24\textwidth]{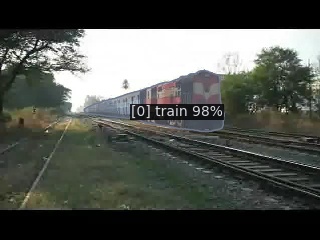}
    \includegraphics[width=0.24\textwidth]{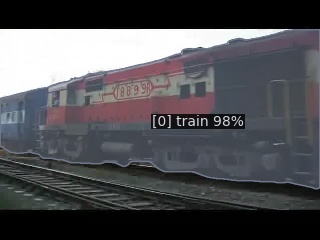}

    \includegraphics[width=0.24\textwidth]{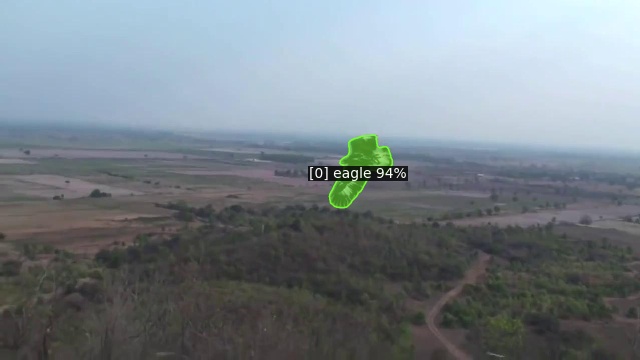}
    \includegraphics[width=0.24\textwidth]{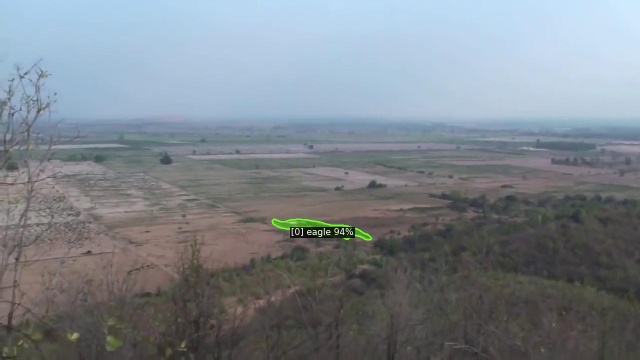}
    \includegraphics[width=0.24\textwidth]{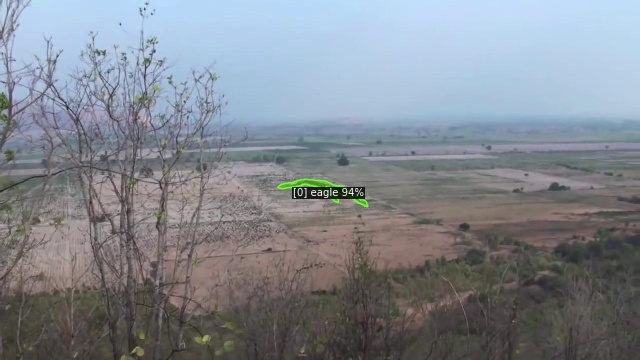}
    \includegraphics[width=0.24\textwidth]{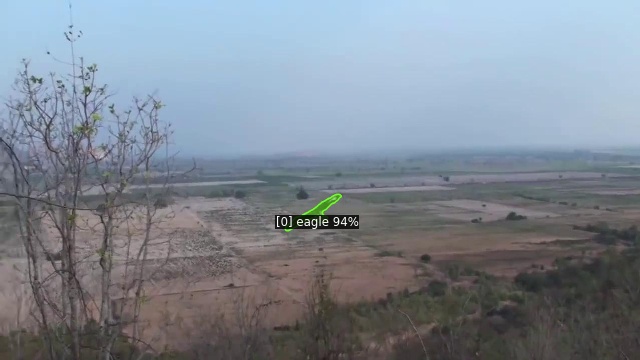}

    \includegraphics[width=0.24\textwidth]{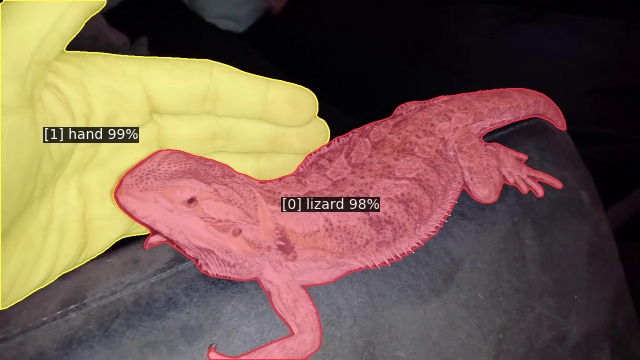}
    \includegraphics[width=0.24\textwidth]{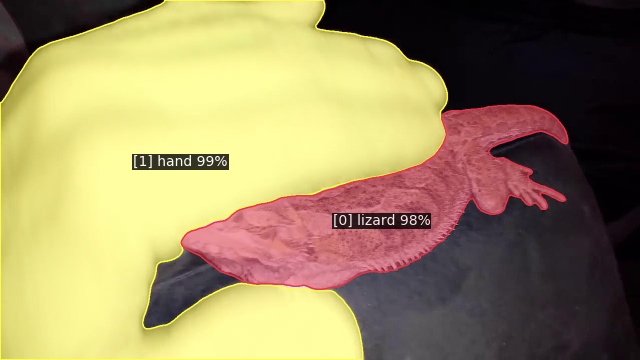}
    \includegraphics[width=0.24\textwidth]{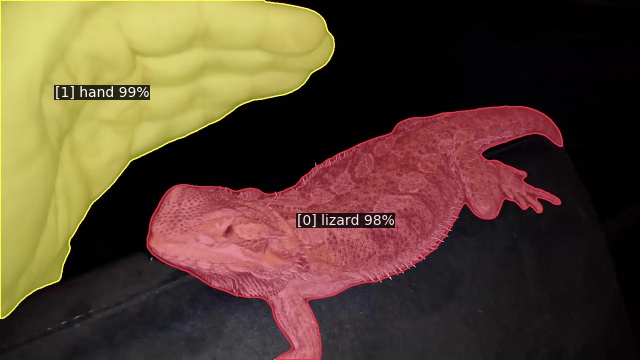}
    \includegraphics[width=0.24\textwidth]{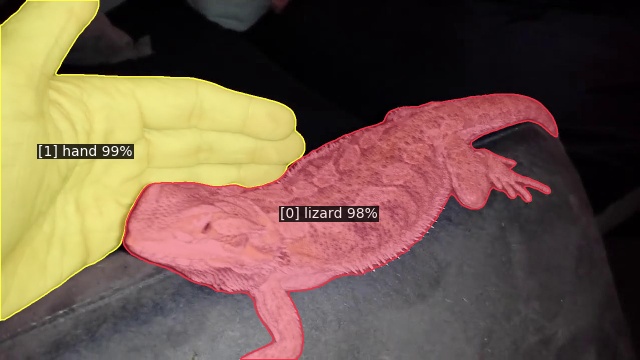}

    \includegraphics[width=0.24\textwidth]{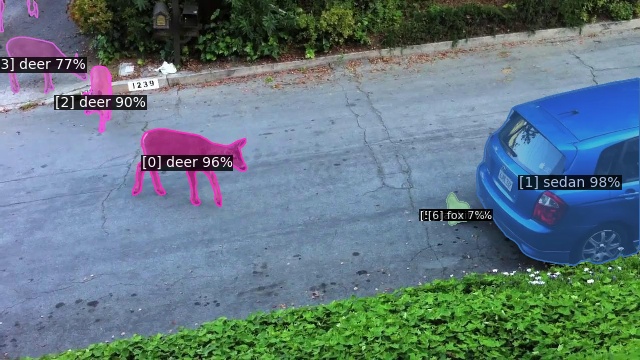}
    \includegraphics[width=0.24\textwidth]{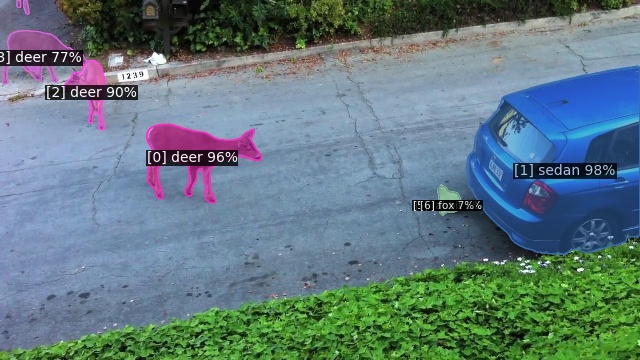}
    \includegraphics[width=0.24\textwidth]{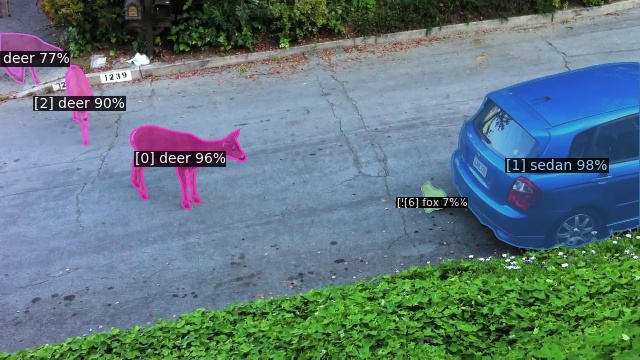}
    \includegraphics[width=0.24\textwidth]{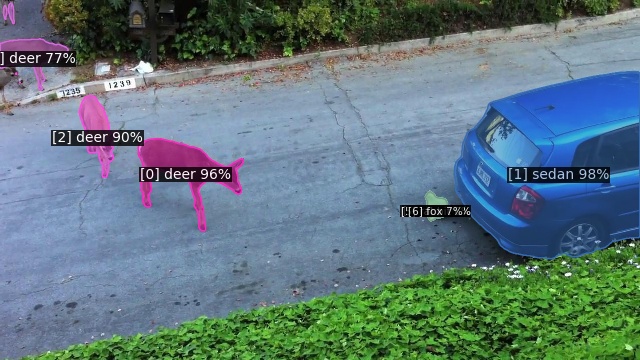}

    \includegraphics[width=0.24\textwidth]{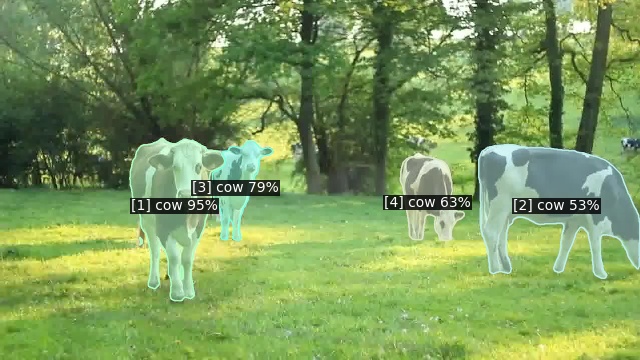}
    \includegraphics[width=0.24\textwidth]{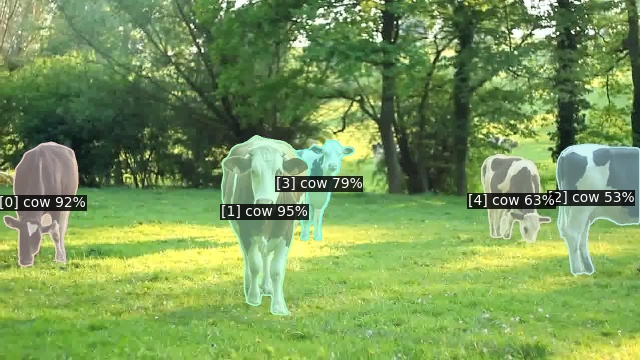}
    \includegraphics[width=0.24\textwidth]{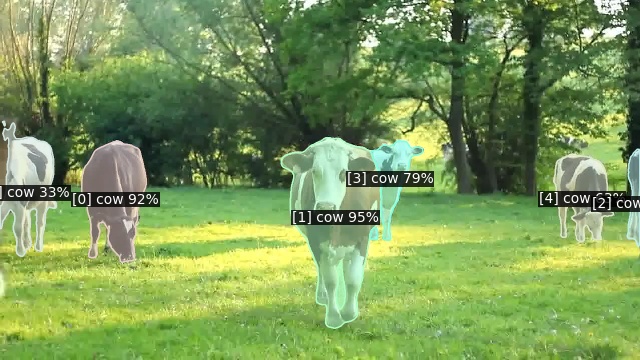}
    \includegraphics[width=0.24\textwidth]{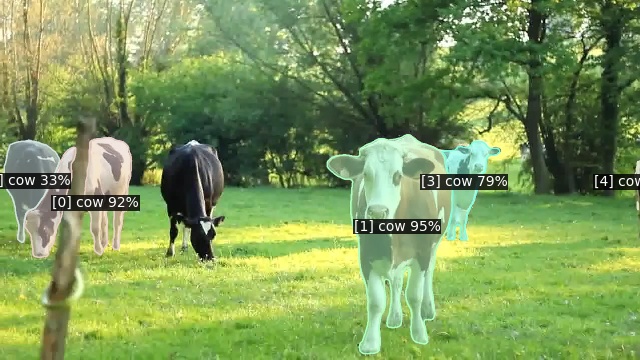}

    \includegraphics[width=0.24\textwidth]{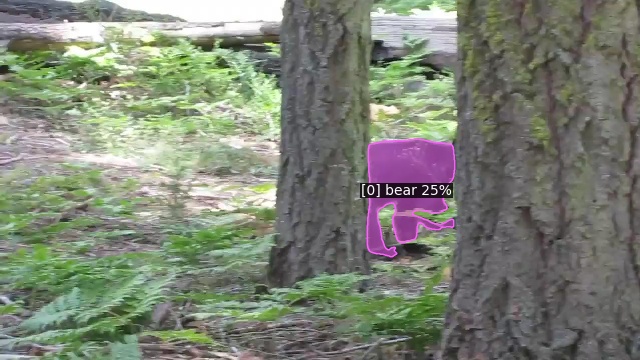}
    \includegraphics[width=0.24\textwidth]{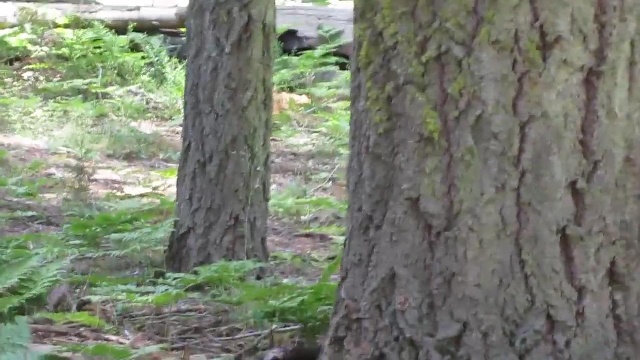}
    \includegraphics[width=0.24\textwidth]{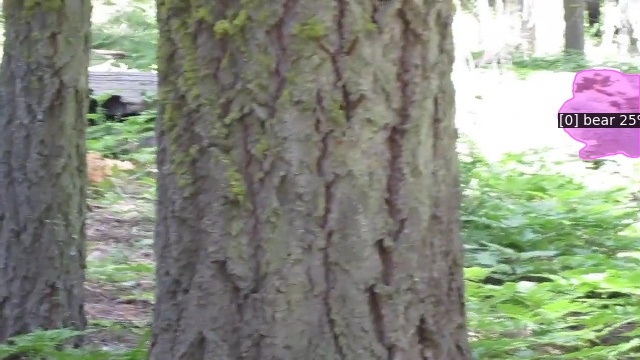}
    \includegraphics[width=0.24\textwidth]{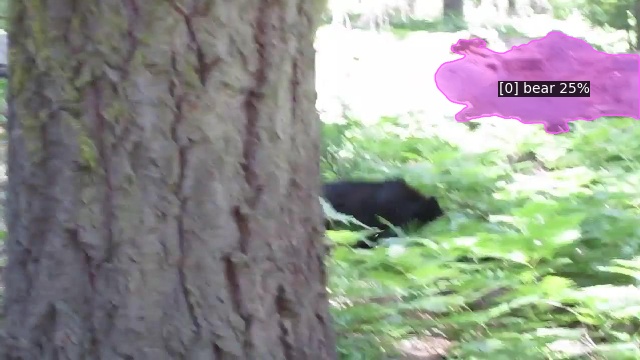}

    \caption{
    Example \textbf{qualitative results} from the \textbf{YouTube-VIS 2019/2021} validation sets.
    We show outputs from our ~\method{} model with the top-performing Swin-L~\cite{SwinTransformer} backbone for 4 frames uniformly selected over the given sequence.
    }
    \label{fig:qualitative_results_ytvis_19_21}
\end{figure*}

\begin{figure*}[t]
    \centering
    \includegraphics[width=0.24\textwidth]{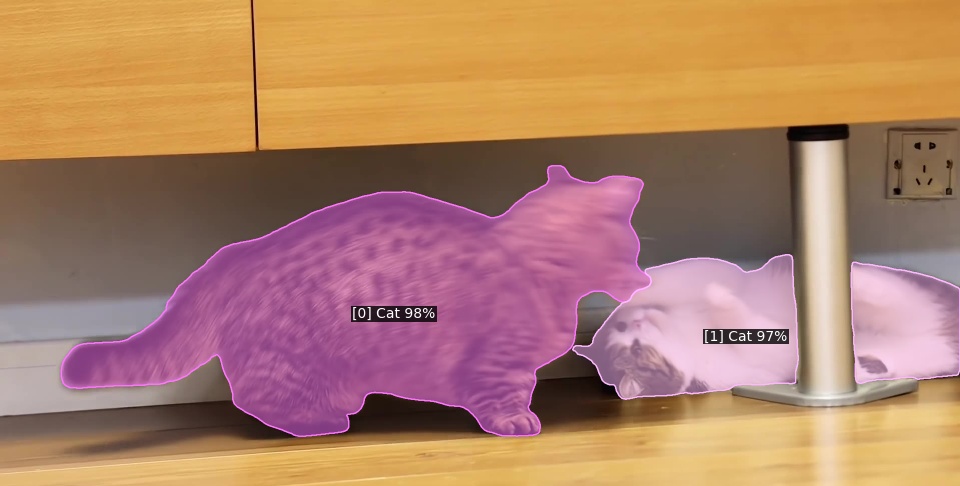}
    \includegraphics[width=0.24\textwidth]{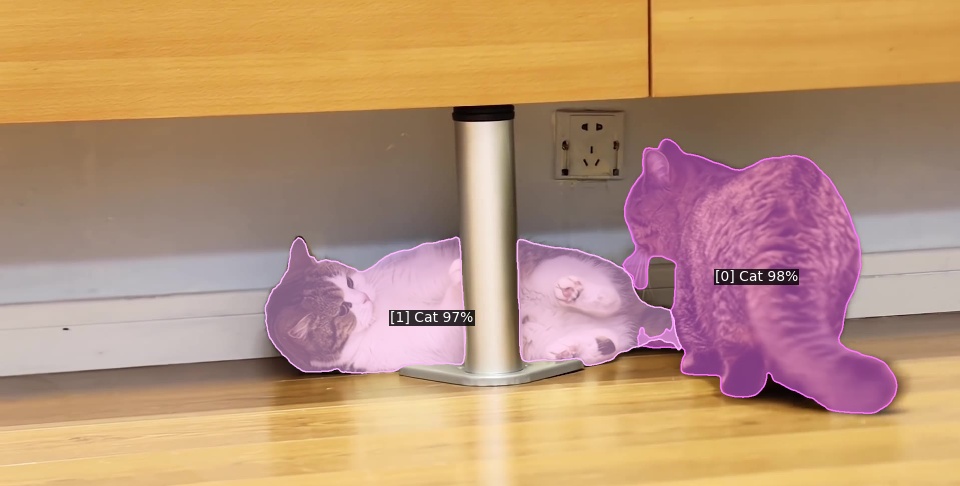}
    \includegraphics[width=0.24\textwidth]{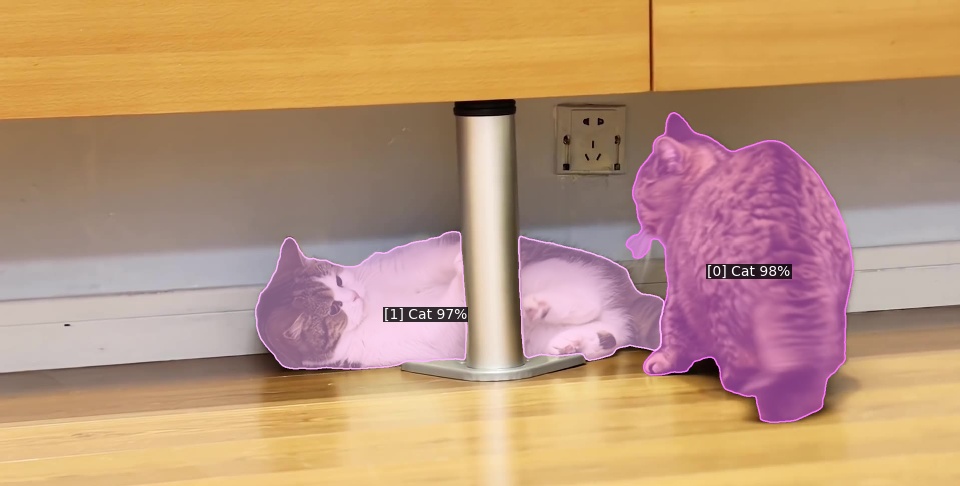}
    \includegraphics[width=0.24\textwidth]{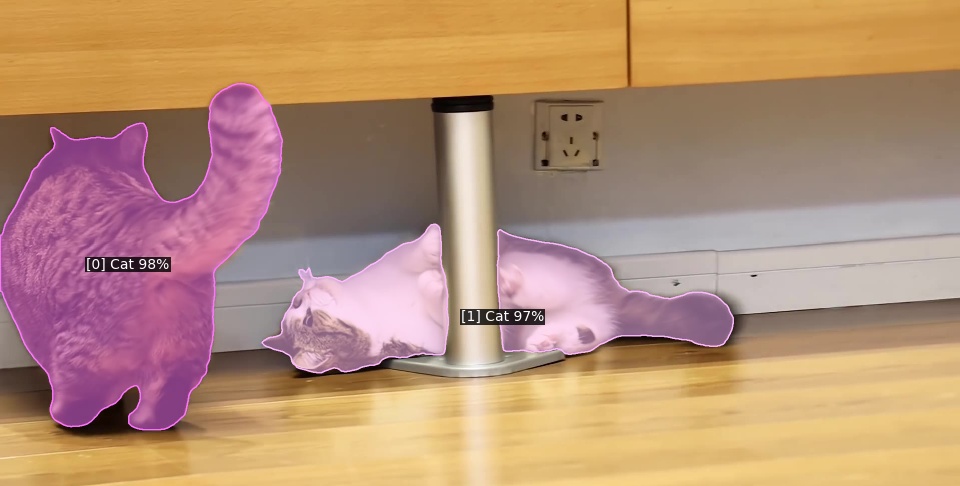}

    \includegraphics[width=0.24\textwidth]{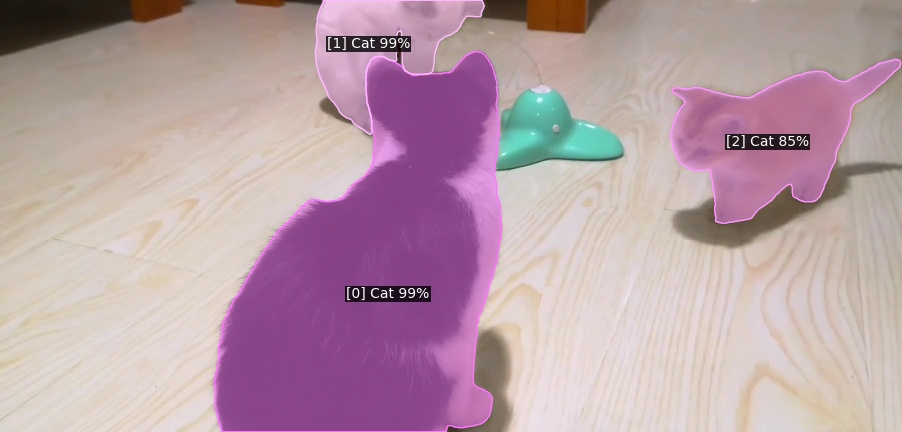}
    \includegraphics[width=0.24\textwidth]{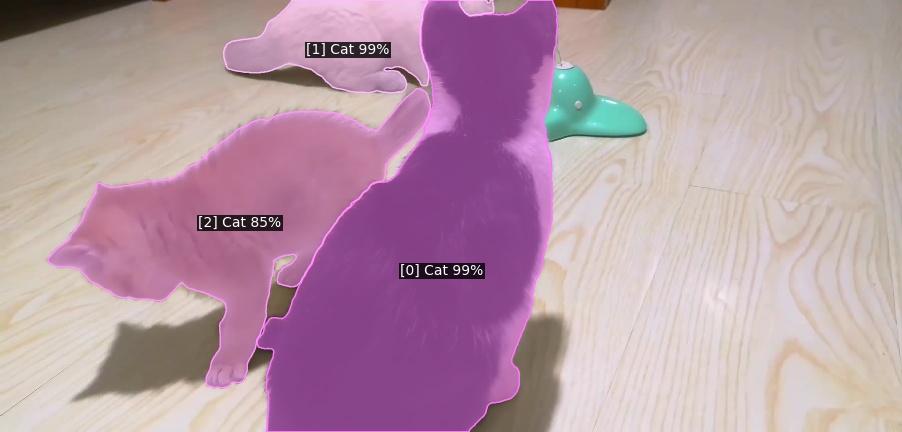}
    \includegraphics[width=0.24\textwidth]{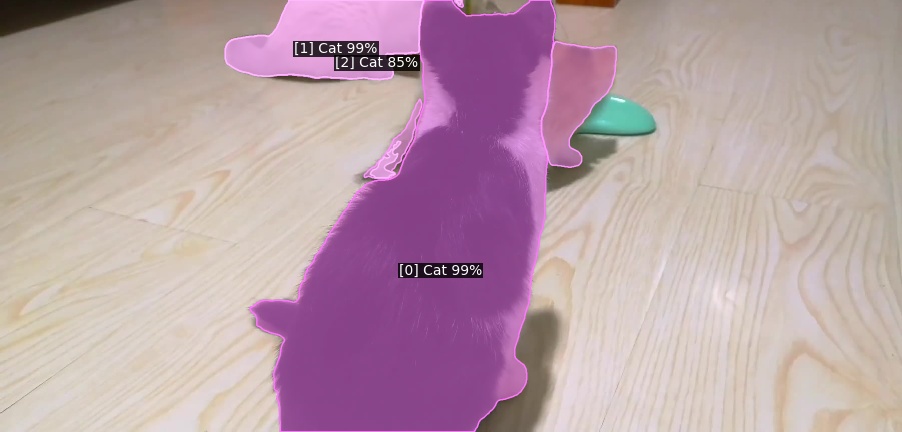}
    \includegraphics[width=0.24\textwidth]{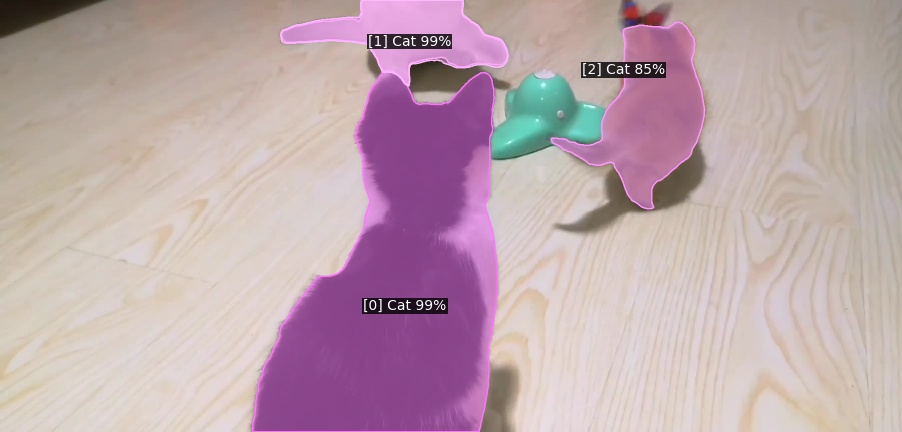}

    \includegraphics[width=0.24\textwidth]{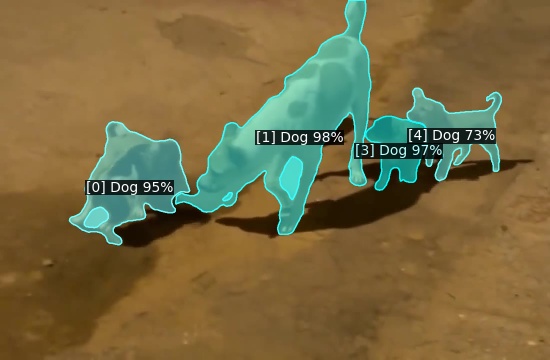}
    \includegraphics[width=0.24\textwidth]{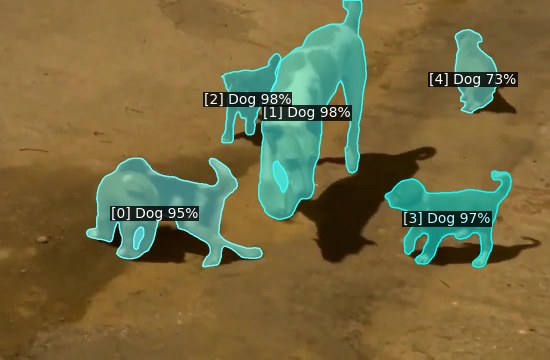}
    \includegraphics[width=0.24\textwidth]{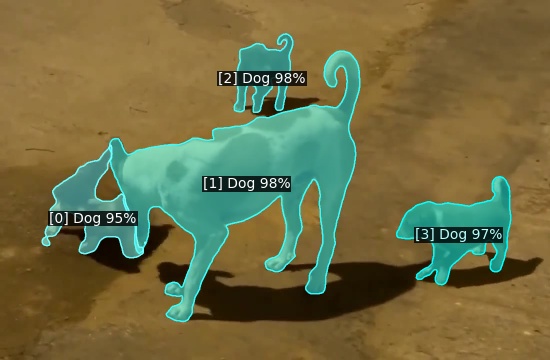}
    \includegraphics[width=0.24\textwidth]{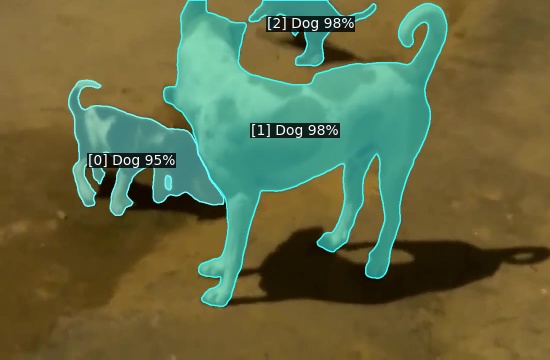}

    \includegraphics[width=0.24\textwidth]{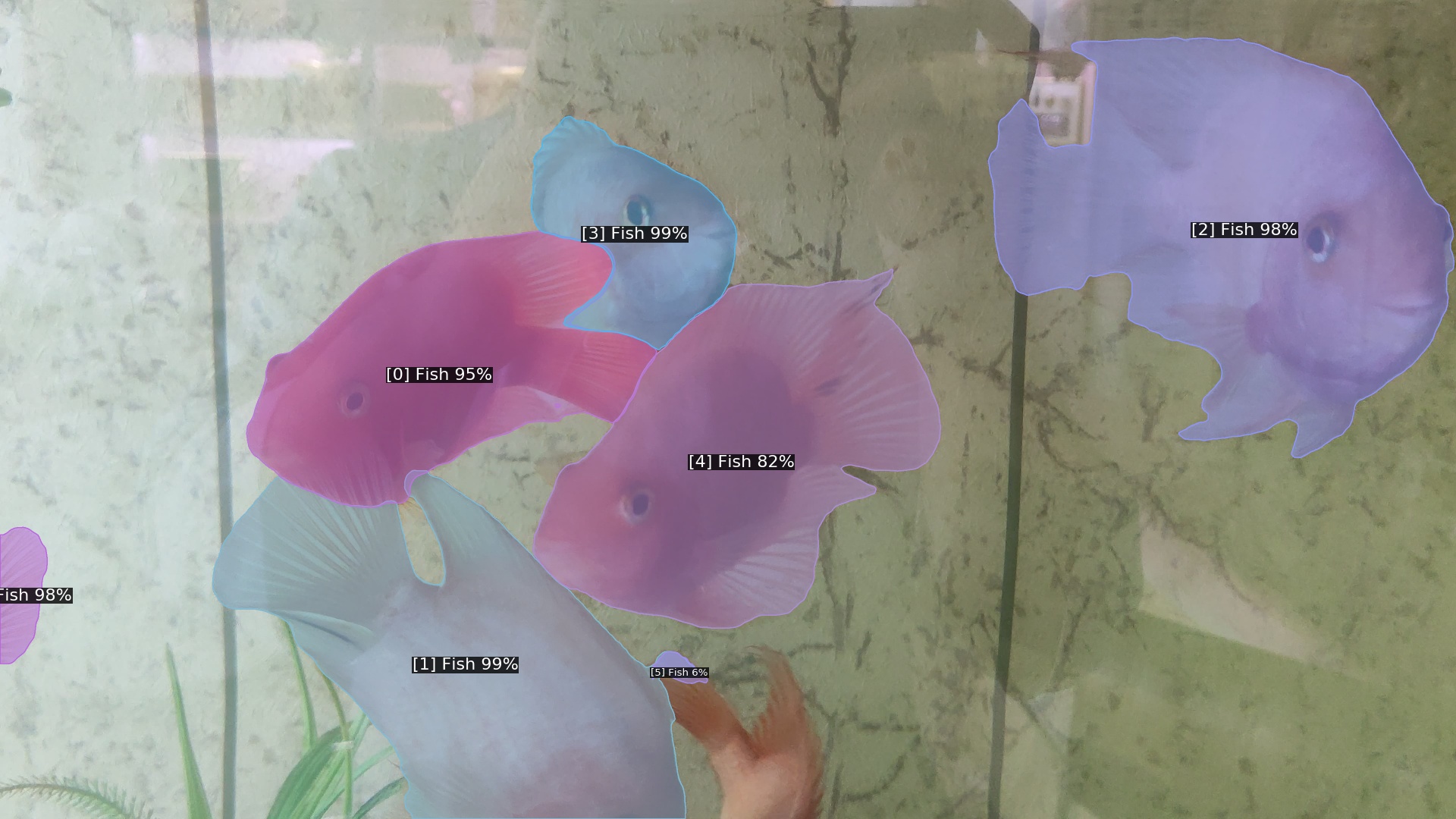}
    \includegraphics[width=0.24\textwidth]{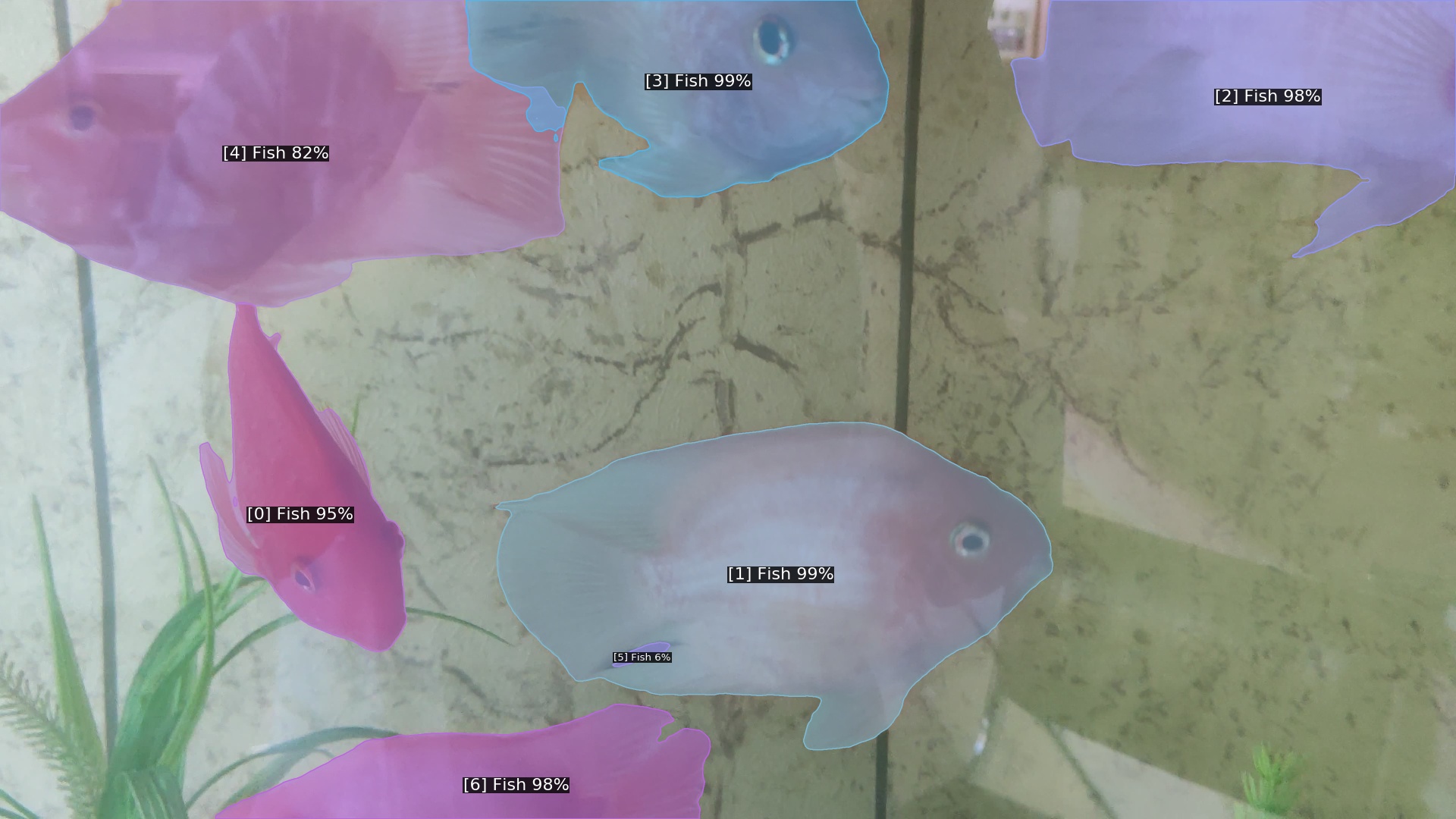}
    \includegraphics[width=0.24\textwidth]{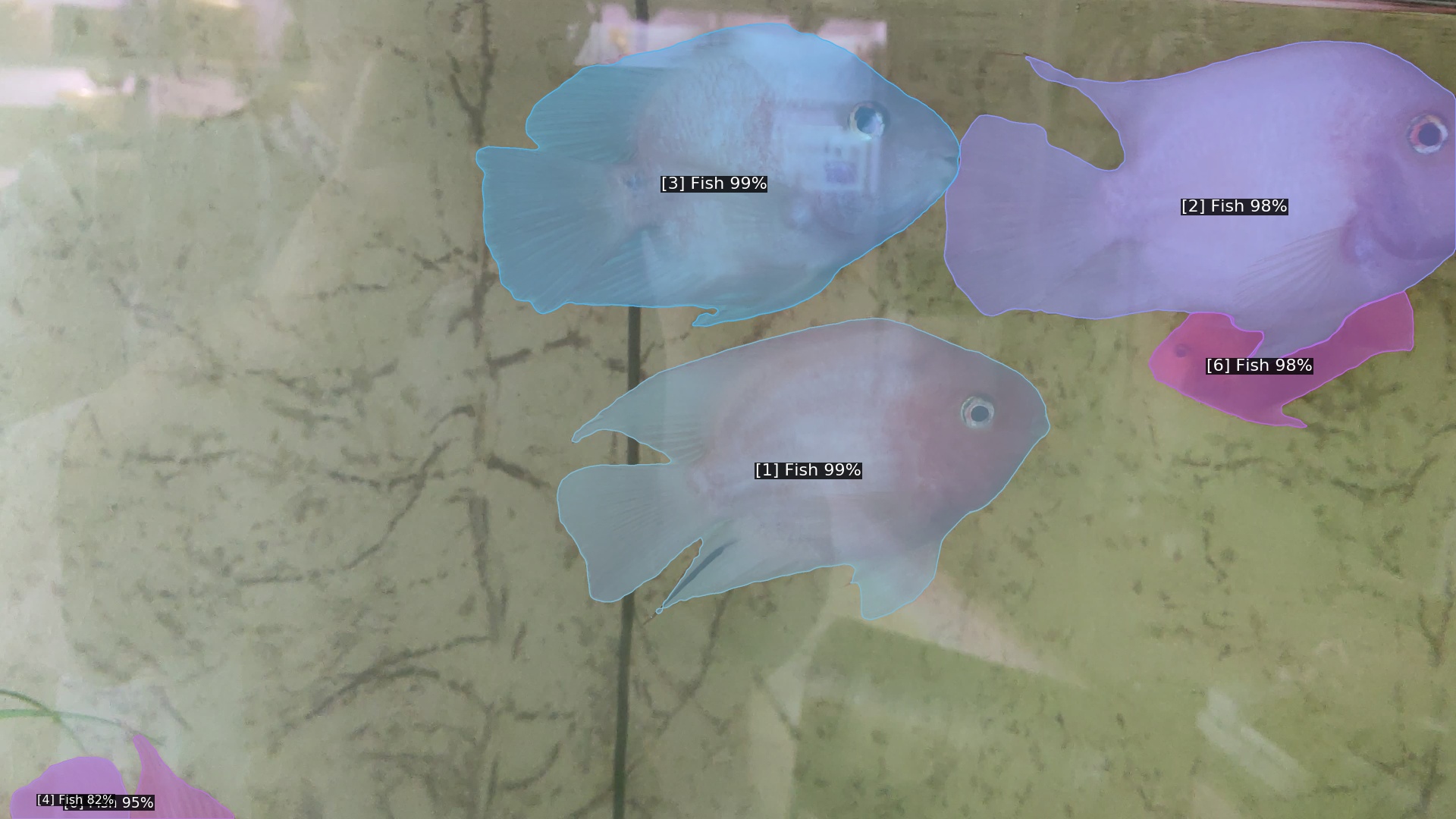}
    \includegraphics[width=0.24\textwidth]{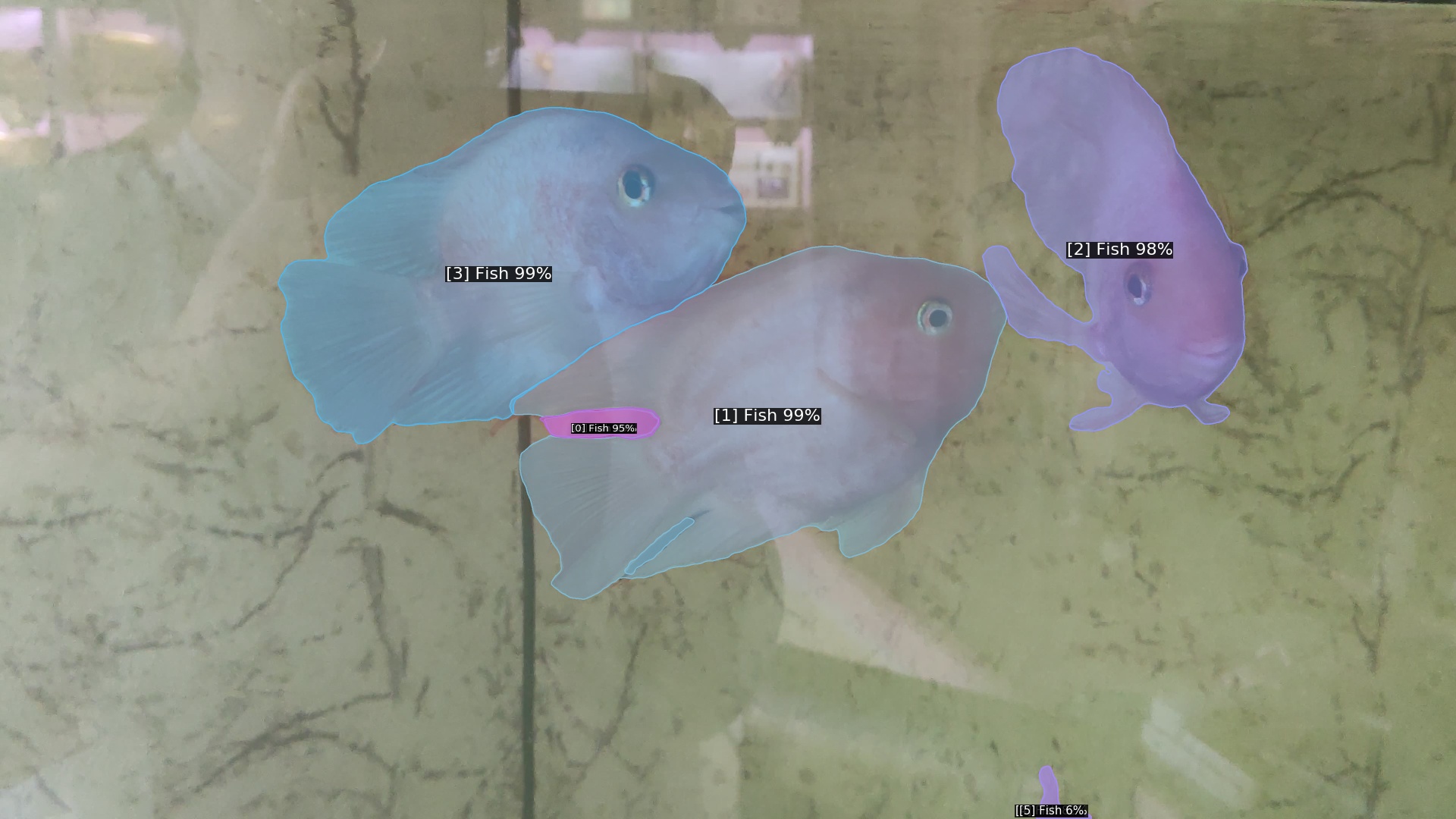}

    \includegraphics[width=0.24\textwidth]{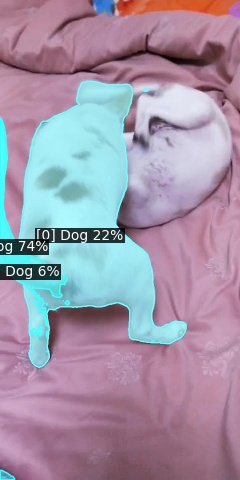}
    \includegraphics[width=0.24\textwidth]{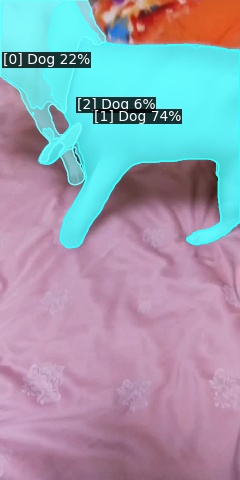}
    \includegraphics[width=0.24\textwidth]{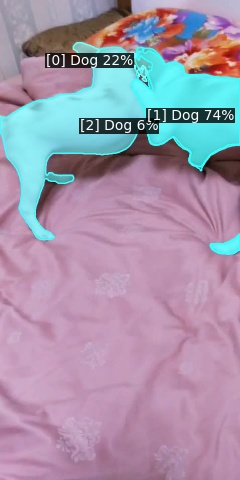}
    \includegraphics[width=0.24\textwidth]{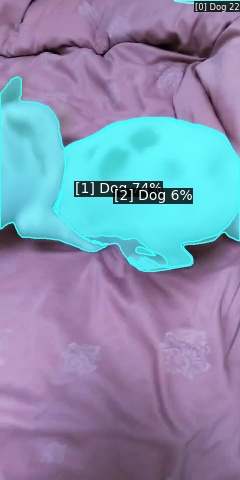}

    \includegraphics[width=0.24\textwidth]{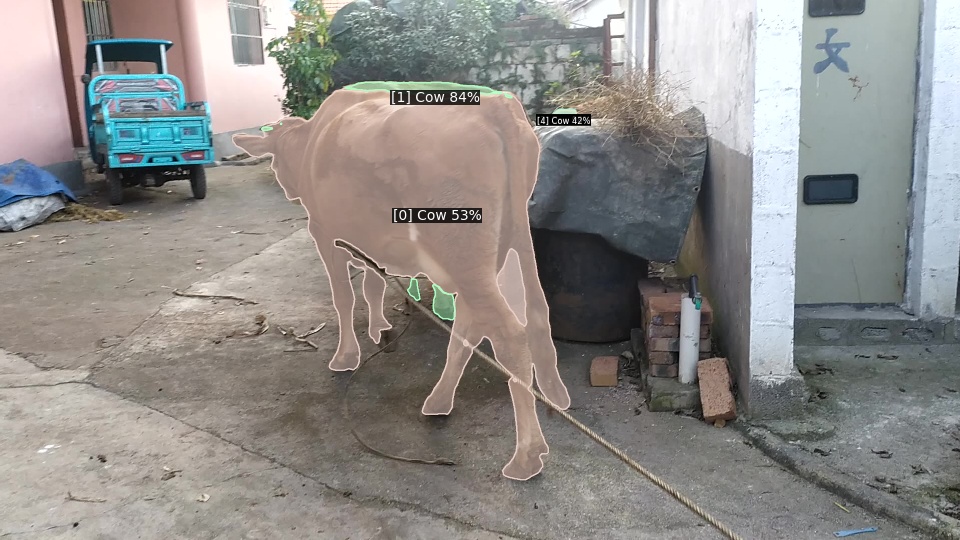}
    \includegraphics[width=0.24\textwidth]{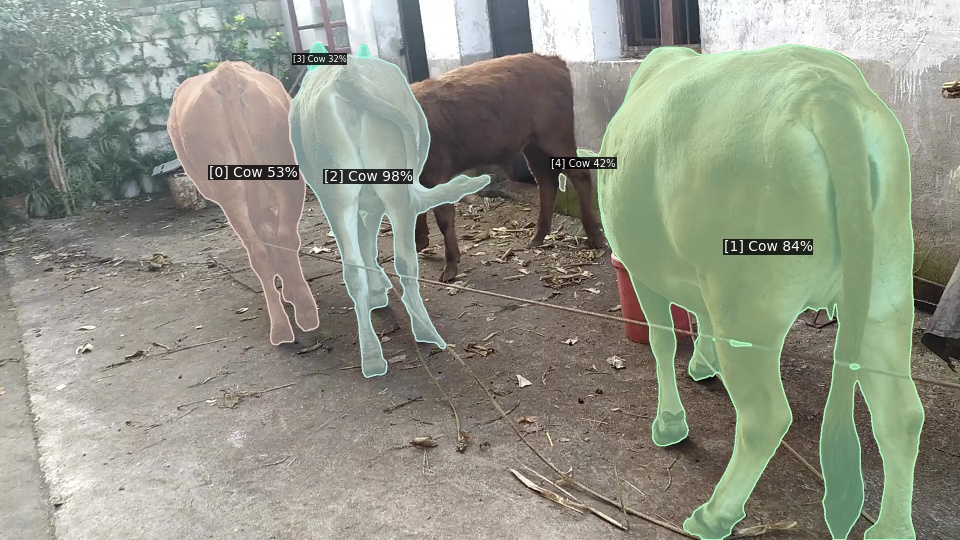}
    \includegraphics[width=0.24\textwidth]{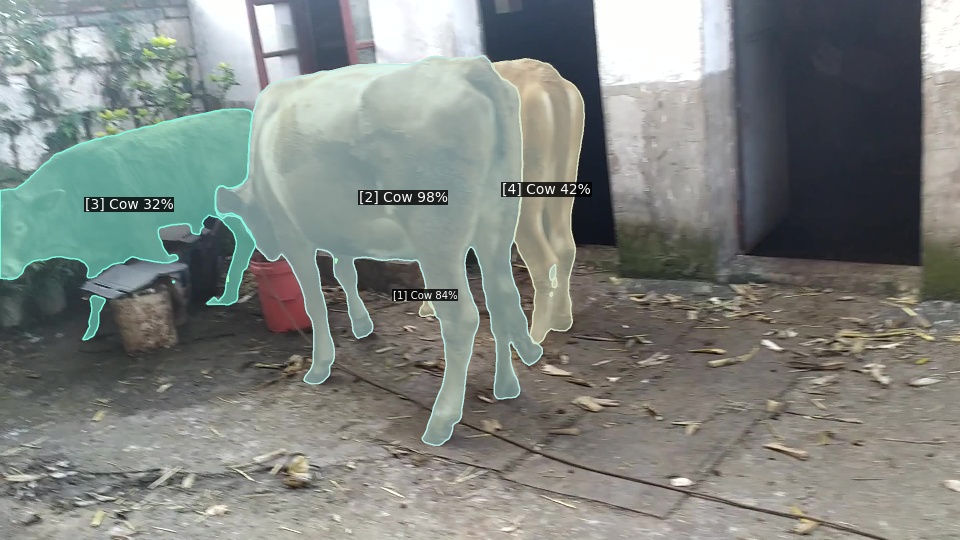}
    \includegraphics[width=0.24\textwidth]{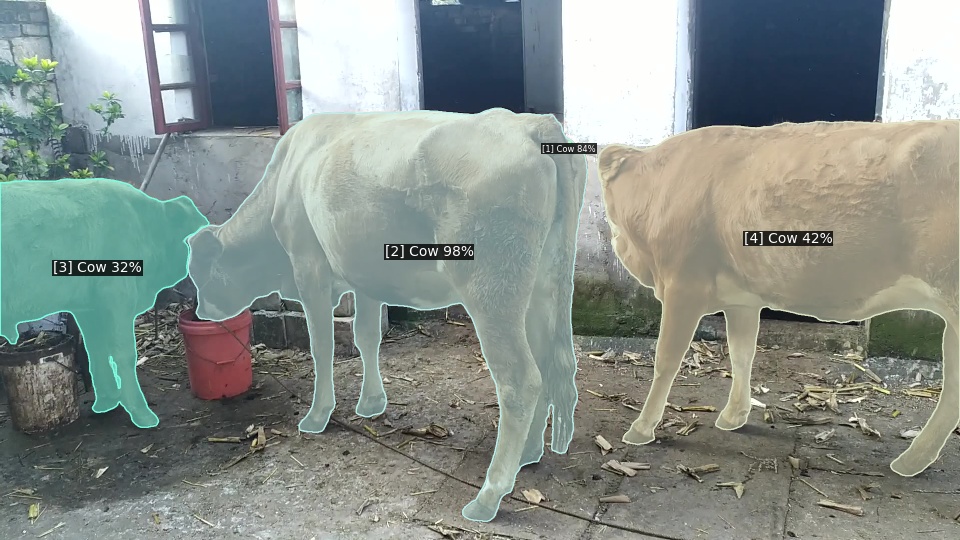}
    
    \caption{
    Example \textbf{qualitative results} from the \textbf{OVIS} validation set.
    We show outputs from our ~\method{} model with the top-performing Swin-L~\cite{SwinTransformer} backbone for 4 frames uniformly selected over the given sequence.
    The model is evaluated with 360 pixels minimum scale at test-time (MST).
    }
    \label{fig:qualitative_results_ovis}
\end{figure*}

The OVIS sequences presented in~\cref{fig:qualitative_results_ovis} represent significantly more challenging VIS scenarios with large object movements, high crowdedness and strong object occlusions.
Our model is able to discriminate several instances even for object classes with similar instance appearance.
In particular, row 4 (fishes) demonstrates strong object tracking capabilities in a scenario challenging even for human observers.
Row 2 (cats) and row 5 (dogs) allow for a comparison of crowded scenes with either small or large object movements, respectively.
As seen in row 2 of~\cref{fig:qualitative_results_ytvis_19_21}, our model is generally able to track large object movements.
However, the combination of crowdedness and object movement in row 5 generates several false positive segmentations.
These are caused by the aforementioned temporal modelling limitations of the pixel decoder.


\begin{table}
    \centering
    \resizebox{0.8\columnwidth}{!}{
        \begin{tabular}{c|cc|c}
            \toprule

            & \multicolumn{2}{c|}{Iterations} &  \\
            Benchmark & LR drops & Total & Top-K  \\
            
            \midrule

            YouTube-VIS 2019~\cite{Yang2019vis}     & 4000, 8000    & 12000 &  10 \\
            YouTube-VIS 2021~\cite{Yang2019vis}     & 6000, 10000   & 14000 &  10 \\
            OVIS~\cite{ovis}                        & 8000, 12000   & 16000 &  30 \\

            \bottomrule
        \end{tabular}
    }
    \caption{
    Summary of \textbf{~\method{} hyperparameters} for several VIS benchmarks.
    In contrast to~\cite{cheng2021mask2former}, we drop the learning rate (LR) twice.
    The OVIS~\cite{ovis} benchmark contains sequences with more than 20 objects, hence we increase the number of top-k predictions at test-time.
    }
    \label{tab:hyperparams_benchmarks}
\end{table}


\clearpage


\end{document}